\documentclass[10pt,twocolumn,letterpaper,hyphens,dvipsnames,table,xcdraw]{article}

\usepackage[pagenumbers]{cvpr} % To force page numbers, e.g. for an arXiv version

\newcolumntype{L}[1]{>{\raggedright\arraybackslash}p{#1}}

\usepackage[utf8]{inputenc} % allow utf-8 input
\usepackage[T1]{fontenc}    % use 8-bit T1 fonts
\usepackage{microtype}      % microtypography

\usepackage{amsmath}
\usepackage{amssymb}        % Loaded only ONCE
\usepackage{amsfonts}       % blackboard math symbols
\usepackage{nicefrac}       % compact symbols for 1/2, etc.
\usepackage{pifont}         % for dingbats like cmark/xmark
\newcommand{\xmark}{\ding{55}}%

\usepackage{booktabs}       % professional-quality tables
\usepackage{multirow}
\usepackage{arydshln}
\usepackage[flushleft]{threeparttable}

\usepackage{graphicx}
\usepackage{multicol}
\usepackage{caption}
\usepackage{wrapfig}
\usepackage{array}
\usepackage{placeins}

\usepackage[most]{tcolorbox}
\tcbuselibrary{listingsutf8}

\usepackage{soul}
\usepackage{enumitem}
\usepackage{tikz}
\usepackage{cuted}
\tcbuselibrary{breakable} 

\definecolor{cvprblue}{rgb}{0.21,0.49,0.74}
\def\paperID{35} % *** Enter the Paper ID here
\def\confName{CVPR}
\def\confYear{2025}

\newtcolorbox{WidePromptBox}[2][]{%
  enhanced,
  breakable,
  colback=gray!10,
  colframe=black!80,
  arc=3mm,
  boxrule=0.8pt,
  fonttitle=\bfseries,
  coltitle=white,
  left=2mm, right=2mm, top=1mm, bottom=1mm,
  sharp corners=southwest,
  before skip=8pt,
  after skip=8pt,
  width=\linewidth,        % full line width in whatever mode you’re in
  title={#2},#1           % user’s title (#2) + any extra keys (#1)
}

\title{Open World Scene Graph Generation using Vision
Language Models}
\author{
    Amartya Dutta\textsuperscript{1*} \quad
    Kazi Sajeed Mehrab\textsuperscript{1‡} \quad
    Medha Sawhney\textsuperscript{1‡} \quad
    Abhilash Neog\textsuperscript{1} \quad
    Mridul Khurana\textsuperscript{1} \\
    Sepideh Fatemi\textsuperscript{1} \quad
    Aanish Pradhan\textsuperscript{1} \quad
    M.~Maruf\textsuperscript{1} \quad
    Ismini Lourentzou\textsuperscript{3†} \\
    Arka Daw\textsuperscript{2†} \quad
    Anuj Karpatne\textsuperscript{1†*} \\
    \textsuperscript{1}Virginia Tech \quad
    \textsuperscript{2}Oak Ridge National Lab \quad
    \textsuperscript{3}University of Illinois Urbana-Champaign \\
}

\begin{document}
\maketitle
\begingroup
\renewcommand\thefootnote{} % Removes the automatic footnote number/symbol
\footnotetext{\textsuperscript{‡} Denotes equal contribution, \textsuperscript{†} denotes equal advising and \textsuperscript{*} denotes the corresponding authors : \texttt{\{amartya, karpatne\}@vt.edu}}
\endgroup
\begin{abstract}
Scene-Graph Generation (SGG) seeks to recognize objects in an image and distill their salient pairwise relationships. Most methods depend on dataset-specific supervision to learn the variety of interactions, restricting their usefulness in open-world settings, involving novel objects and/or relations. Even methods that leverage large Vision Language Models (VLMs) typically require benchmark-specific fine-tuning. We introduce Open-World SGG, a training-free, efficient, model-agnostic framework that taps directly into the pretrained knowledge of VLMs to produce scene graphs with zero additional learning. Casting SGG as a zero-shot structured-reasoning problem, our method combines multimodal prompting, embedding alignment, and a lightweight pair-refinement strategy, enabling inference over unseen object vocabularies and relation sets. To assess this setting, we formalize an Open-World evaluation protocol that measures performance when no SGG-specific data have been observed either in terms of objects and relations. Experiments on Visual Genome, Open Images V6, and the Panoptic Scene Graph (PSG) dataset demonstrate the capacity of pretrained VLMs to perform relational understanding without task-level training.
\end{abstract}

\section{Introduction}

% The primary goal of Scene Graph Generation (SGG) is to convert an image into a graph-based representation that captures the visual content through object entities and the semantic relationships between them. This structured and interpretable format serves as a bridge between visual perception and high-level reasoning, making it a valuable intermediate representation. Scene graphs have proven effective as a foundation for a variety of downstream tasks, including image captioning, visual question answering (VQA), referring expression generation, and broader visual reasoning applications \cite{hildebrandt2020scene,hudson2019learning,hudson2019gqa,shi2019explainable,teney2017graph,johnson2015image,yang2019cross,yang2019auto,yang2022reformer}.

Scene Graph Generation (SGG) aims to convert an image into a structured graph, where nodes correspond to object entities and edges capture the semantic relationships between them. This intermediate representation enables structured reasoning over visual content and has been shown to benefit a range of downstream tasks, including image captioning, visual question answering, and referring expression generation \cite{johnson2015image,teney2017graph,hudson2019gqa,yang2019cross}. Achieving accurate SGG requires a strong understanding of both visual appearance and the contextual semantics that govern interactions between objects.

% Traditional SGG methods rely heavily on fully supervised learning pipelines trained on annotated scene graph datasets such as Visual Genome. A considerable body of work has addressed challenges such as the long-tail distribution of object and relationship classes \cite{xxx}. While these models have achieved strong performance under constrained settings, their real-world applicability remains limited due to the high cost of annotation and their reliance on a fixed, closed vocabulary.

Traditional approaches to SGG are predominantly supervised, trained on datasets like Visual Genome that provide dense annotations of object and relationship triplets. While these models have made significant progress, they are fundamentally constrained by the scope and vocabulary of the annotated data. Annotation is labor-intensive and often biased, and the long-tail distribution of object and predicate classes further hampers generalization to complex, real-world imagery \cite{zellers2018neural,lu2016visual}. To address these limitations, recent efforts have explored Open-Vocabulary SGG (OV-SGG), where the goal is to predict objects or relations that were not seen during training. This includes tasks such as Open-Vocabulary Object Detection (OVD) -- detecting unseen object categories - and Open-Vocabulary Relationships (OVR) -- predicting unseen predicates between known object pairs \cite{gu2021open}. More recent formulations consider fully open-world settings where objects or predicates may be novel at inference time \cite{ovsgtr,rahp}. However, even these models rely on fine-tuning or auxiliary training stages, limiting their adaptability and requiring curated data.

% To overcome these limitations, recent research has focused on Open-Vocabulary Scene Graph Generation (OV-SGG), which enables generalization to novel object or predicate categories not seen during training. This line of work typically considers two sub-tasks: Open-Vocabulary Detection (OvD) \cite{xxx}, which predicts known predicates between previously unseen object categories, and Open-Vocabulary Relationships (OvR) \cite{xxx}, which classifies novel predicates between known object categories. More recent efforts have proposed settings that combine both OvD and OvR, where both objects and relationships are novel at test time \cite{xxx}. However, existing methods in this space still depend on task-specific training or additional supervision, limiting their flexibility and scalability in truly open-world environments. Thus, building a general-purpose, training-free SGG system capable of operating in a fully open-vocabulary setting remains an open challenge. 

Given the rapid advancements of Vision-Language Models (VLMs) trained on massive image-text corpora \cite{qwen-vl, molmo, llava, gpt4o, team2024gemini}, a natural question arises: \textit{Can VLMs enable zero-shot scene graph generation without requiring task specific training?}  These models demonstrate strong visual and language generalization, and recent work has proposed reformulating SGG sub-tasks -- particularly predicate classification -- as image-text matching problems. Yet, most of these approaches still incorporate dataset-specific modules or rely on expensive pairwise inference pipelines, therefore not evaluating VLMs on a truly open-world setting. 

Despite the growing interest in leveraging VLMs for SGG, progress in evaluating these models for SGG has been hindered by several key limitations. \textit{First}, there is a lack of standardized baselines for open-world SGG, making it difficult to assess how well models generalize to unseen categories. \textit{Second}, there is no established methodology for prompting VLMs to generate scene graphs in a way that is both effective and comparable with existing methods. \textit{Third}, the open-ended nature of VLM outputs makes it nontrivial to elicit structured predictions like subject–predicate–object triplets, which would enable their outputs to be quantitatively evaluated with popular SGG datasets, like Visual Genome \cite{visualgenome}, Panoptic Scene Graph (PSG) \cite{psg} and OpenImage (OI) \cite{oiv6}.

To bridge this gap, we present Open World SGG (OwSGG) an end-to-end, model-agnostic framework for zero-shot scene graph generation using pretrained VLMs. Our method combines multimodal prompting, embedding alignment, and a lightweight pair-refinement strategy to transform raw VLM outputs into structured scene graphs that are compatible with existing evaluation protocols, enabling quantitative benchmarking without requiring any task-specific training. Using this framework, we conduct a comprehensive evaluation of two popular VLMs, LLaVa-next \cite{llava} and Qwen2-VL~\cite{qwen2-vl} -- across a range of settings -- including closed vocabulary, open-vocabulary objects and open-vocabulary relationships. We further introduce a fully open-world case. While we do not obtain the best results in the closed-world setting, we find that VLMs, despite no access to task-specific training, can match or even surpass these models in certain open-world cases. To encourage further research, we introduce an open-world baseline that isolates the performance of VLMs on novel object AND novel relation pairs, providing a new point of comparison for future methods. Our findings highlight the potential of pretrained vision-language models for scalable scene graph understanding and underscore the need for new methods and benchmarks tailored to the open-world setting.

\section{Related Works}

\textbf{Scene Graph Generation} introduced in \cite{visualgenome}, aims to localize, classify and predict relationships between entities in images, enabling structured visual understanding. Early works in SGG \citep{johnson2015image, xu2017scene} predict pairwise relationships between objects to construct graphs and demonstrate various scene graph applications, such as Image Captioning and VQA \citep{hildebrandt2020scene, hudson2019gqa}. \citep{zellers2018neural} improve SGG by higher-order “motifs’’ in the object–predicate–object statistics, yielding large gains.
However, these methods rely on fully supervised training with labeled scene graph data, which is expensive to annotate, difficult to scale to the wide variety of objects and relationships that can exist in natural scenes, and can suffer from data imbalance. To mitigate these, weakly and semi-supervised approaches \cite{kim2024llm4sgg} and imbalanced learning strategies \cite{cong2023reltr} have been explored, but they remain restricted to a closed vocabulary of relationships seen during training. Recent work has introduced open-vocabulary SGG (OV-SGG), in two primary settings: Open-vocabulary Detection (OvD), which predicts known predicates between unseen object categories \cite{ovsgtr, he2022towards, zareian2021open, vs3}, and Open-vocabulary Relationships (OvR), which classifies unseen predicates between known object categories \cite{ovsgtr, zareian2021open}. A more recent line of work addresses the combined OvD+OvR setting, though still relying on task-specific supervision \cite{ovsgtr, vs3}. Tackling a truly open-world, training-free SGG setting -- where models generate scene graphs without any fine-tuning or supervision was previously infeasible due to limitations in model capabilities. However, recent progress in zero-shot object detection, language modeling, and vision-language models (VLMs) now makes it possible to evaluate whether such models can construct scene graphs without any fine-tuning, enabling a truly open-world and training-free approach to structured visual understanding.

\textbf{Vision-Language Models (VLMs) for Scene Graph Generation} have become a popular choice for SGG in recent times, given the large amount of pre-trained knowledge VLMs have. Early VLMs such as \cite{clip, li2022blip, li2023blip} have shown promising performance over image and text modalities. This encouraged the development of the more recent VLMs \cite{qwen-vl, molmo, llava, gpt4o, team2024gemini}
% [LLava,Molmo,Qwen2vl,Gemini, GPT4o] 
which have shown promising performance and have become popular as foundation models capable of several Vision and Language tasks. \cite{ovsgtr, from-pixels-to-graphs} have used pre-trained VLMs for predicting Open Vocabualry SG relations. While these methods do utilise the knowledge of VLMs, they still include some dataset or task-specific training \cite{llavaspace,gpt4sgg}. Besides, just using the language priors of VLMs for predicting unseen relationships, \cite{prism, rahp} also use VLMs and LLMs for object pair refinement before predicting relationships between them. This is an expensive process ranging in the $\mathcal{O}(n^2)$ operations. We address these limitations by proposing a lightweight pair refinement module in our evaluation framework that enables the use of off-the-shelf VLMs in a training free and lightweight manner to generate and evaluate SGs in an Open-World setting.

\section{Open-world Scene Graph Generation (Ow-SGG)}

In this section, we present our {proposed framework for Open-world Scene Graph Generation (SGG) using Vision-Language Models (VLMs)}. We begin by introducing some background of SGG and notations used throughout this work. We then {introduce the taxonomy of problem setups within the Open-world SGG (Ow-SGG) paradigm, delineating the novel challenges for each one of them}. Finally, we describe our proposed framework for leveraging VLMs for SGG, highlighting how such models can be used in open-world settings.

% \ak{\st{Introduce an overview of this section. We are going to introduce the "Background and Notations", "Proposed Framework for using VLMs for Scene Graph Generation (SGG)", and " Categorizing Problems in Open World SGG"}}

% -- introduce a new task OW (Open world) SGG.  Triplets where **both** predicate **and** subject/object are novel. Only triplets composed entirely of novel elements are included.OW ⊂ OVDR: OW is a subset of OVDR (more restrictive on novelty).

\subsection{Background and Notations}
The goal of Scene Graph Generation (SGG) is to construct a structured graph-based representation of an image’s visual content, termed as a scene graph. Formally, a scene graph is defined as \(\mathcal{G} = (\mathcal{V}, \mathcal{E})\), where \(\mathcal{V} = \{\mathbf{v_i}\}_{i=1}^{N}\) is the set of nodes representing objects, and \(\mathcal{E} = \{\mathbf{e_{ij}}\} \subseteq \mathcal{V} \times \mathcal{V} \times \mathcal{R}\) is the set of directed edges encoding pairwise relationships between objects, $r_{ij} \in \mathcal{R}$, where $\mathcal{R}$ is a fixed set of relation (predicate) classes.

Every object \(\mathbf{v_i}\) = $(\mathbf{b_i}, o_i)$ is associated with a class label \(o_i \in \mathcal{O}\), where \(\mathcal{O}\) is the predefined set of object categories, and a bounding box \(\mathbf{b_i} \in \mathbb{R}^4\), which specifies its spatial location within the image. Each directed edge \(\mathbf{e_{ij}} = (\mathbf{v_i, v_j}, r_{ij})\) represents a relationship from object \(\mathbf{v_i}\) to object \(\mathbf{v_j}\), labeled by \(r_{ij} \in \mathcal{R}\). Based on the amount of information available for generating an image's scene graph, there are two problem formulations of SGG that we consider in our work.

% \vspace{0.5em}
\noindent\textbf{Predicate Classification (PredCls).} In this problem formulation, we are given an image \(I \in \mathbb{R}^{H \times W \times C}\) and the set of objects, $\mathcal{V}$, present in the image. The goal of PredCls is then to predict $\mathcal{E}$, i.e., the object pairs $(\mathbf{v_i},\mathbf{v_j})$ present in the scene graph and its predicate class \(r_{ij}\). Note that PredCls requires ground-truth knowledge of the set of objects $\mathcal{V}$ present in the image and hence can be considered as a restricted problem setting of SGG.
% \(\mathbf{v_i}\) and  \(\mathbf{v_j}\) . \ak{\st{should this not be simply $\mathcal{E}$ given $\mathcal{V}$ and $I$?}}

\noindent\textbf{Scene Graph Detection (SGDet).} Given an image $I$, the goal of the SGDet is to detect both the set of objects, $\mathcal{V}$, and the semantic relationships, $\mathcal{E}$, between all object pairs in the scene graph. The output is a set of triplets, $\mathcal{E} = \left\{ (\mathbf{v_i}, \mathbf{v_j}, r_{ij}) \mid \mathbf{i} \ne \mathbf{j},\ r_{ij} \in \mathcal{R},\ \mathbf{v_i}, \mathbf{v_j} \in \mathcal{V} \right\}$, that compactly represent the structured content of the image. SGDet explores a more general problem setting of SGG than PredCls.
% \ak{\st{Is $\mathcal{T}$ the same as $\mathcal{E}$? Can we use the same terminology?}}
% \ak{\st{this seems similar to the last paragraph. Shouldd we say SGG and SGDet are different?}}
% \ak{\st{should this not be simply $\mathcal{V}$ and $\mathcal{E}$ given $I$?}}

% \vspace{0.5em}
% \noindent\textbf{Scene Graph Detection (SGDet).} \ak{this seems similar to the last paragraph. Shouldd we say SGG and SGDet are different?} In the most general setting, the model is given only the raw image \(I\), and must detect
% $\{(\mathbf{b_i}, o_i)\}_{i=1}^{N}$ and $\{(\mathbf{i, j}, r_{ij}) \mid \mathbf{i} \ne \mathbf{j},\ r_{ij} \in \mathcal{R}\}$, where both objects and their pairwise relationships must be inferred from scratch. \ak{should this not be simply $\mathcal{V}$ and $\mathcal{E}$ given $I$?}

% The final output of any SGG method is the full scene graph \(\mathcal{G} = (\mathcal{V}, \mathcal{E})\), reconstructed from the detected triplets, capturing the complex semantics and spatial structure of the visual scene.

\subsection{{Open-world Taxonomy}}
We consider a range of task settings in SGG to benchmark the performance of new and existing methods in open-world settings. These tasks are defined by the novelty of triplet components \((o_i, o_j, r_{ij}) \in \mathcal{E}_{\text{test}}\) observed during testing, with respect to the set of training triplets $\mathcal{E}_{\text{train}}$ in terms of novel objects, novel relations, or both.
% \ak{\st{This should be moved to Section 3 as it is also a contribution that will be  ignored as a known experiment setup if left here. For each OW setup, we should explain what previous works have explored it.}}

% \textbf{CS (Closed-Set).} All triplets whose objects and predicates are within the training vocabulary. No filtering is applied.
  
% \textbf{ZS (Zero-Shot)}. Triplets where the object–predicate–object combination has never appeared in the training set, although individual objects and predicates have been seen.

% \textbf{OVR (Open-Vocabulary Relations)}. Triplets in which the predicate is novel (not seen during training). Objects may be seen or unseen.
  
% \textbf{OVD (Open-Vocabulary Object Detections)}. Triplets in which the objects are novel (not seen during training). Predicates may be seen or unseen.

% \textbf{OVD+R (Open-Vocabulary Detection + Relations)} Union filter: triplets where predicate OR subject/object is novel. Allows unseen objects and predicates.

% \textbf{OW (Open World)}. Triplets where both predicate AND subject/object are novel. Only triplets composed entirely of novel elements are included.OW $\subset$ OVDR: OW is a subset of OVDR (more restrictive on novelty).

\textbf{Close Vocabulary (CS).} All triplets in this setting consist of object pairs and predicate combinations that have been observed during training. Formally, for every \((o_i, o_j, r_{ij})\) in $\mathcal{E}_{\text{test}}$, we have seen a different instance of \((o_i, o_j, r_{ij})\) in $\mathcal{E}_{\text{train}}$.

\textbf{Zero-Shot (ZS).} In this setting, the full triplet combination \((o_i, o_j, r_{ij}) \in \mathcal{E}_{\text{test}}\) has not been seen during training, but individual components are known, i.e., while \((o_i, o_j, r_{ij}) \notin \mathcal{E}_{\text{train}}\), \(o_i, o_j \in \mathcal{O}_{\text{train}}\) and \(r_{ij} \in \mathcal{R}_{\text{train}}\), where \(\mathcal{O}_{\text{train}}\) and \(\mathcal{R}_{\text{train}}\) denote the set of observed objects and relation classes.

\textbf{Open-Vocabulary Relations (OVR).} This open-world task setting of SGG explores the scenario where object classes are known but we are interested in detecting predicate classes that we have never seen before during training. Formally, for every $(o_i, o_j, r_{ij}) \in \mathcal{E}_{\text{test}}$, while \(o_i, o_j \in \mathcal{O}_{\text{train}}\), \(r_{ij} \notin \mathcal{R}_{\text{train}}\).

\textbf{Open-Vocabulary Detections (OVD).} Here we consider the scenario where object classes are novel but the relation classes are known, i.e., for every $(o_i, o_j, r_{ij}) \in \mathcal{E}_{\text{test}}$, while \(r_{ij} \in \mathcal{R}_{\text{train}}\), \(o_i, o_j \notin \mathcal{O}_{\text{train}}\).

\textbf{Open-Vocabulary Detections + Relations (OVD+R).} This open-world task setting considers a union of OVD and OVR where either the object classes are novel OR the predicate classes are novel. Formally, for every $(o_i, o_j, r_{ij}) \in \mathcal{E}_{\text{test}}$, we have
\(
(o_i, o_j \notin \mathcal{O}_{\text{train}}) \lor (r_{ij} \notin \mathcal{R}_{\text{train}}).
\)

\textbf{Open World (OW).} A special case of OVD+R setting is when both object classes AND predicate classes are novel. This represents the most challenging setup that we refer to as the strictly Open World (OW) setting, formally defined as 
\(
\forall ~(o_i, o_j, r_{ij}) \in \mathcal{E}_{\text{test}},~ (o_i, o_j \notin \mathcal{O}_{\text{train}})  \land (r_{ij} \notin \mathcal{R}_{\text{train}}).
\)

\subsection{{Proposed Framework for Using VLMs for Ow-SGG}}

\begin{figure*}[!t]  % or [!b] ; starred env = across both columns
  \centering
  \includegraphics[width=0.85\textwidth]{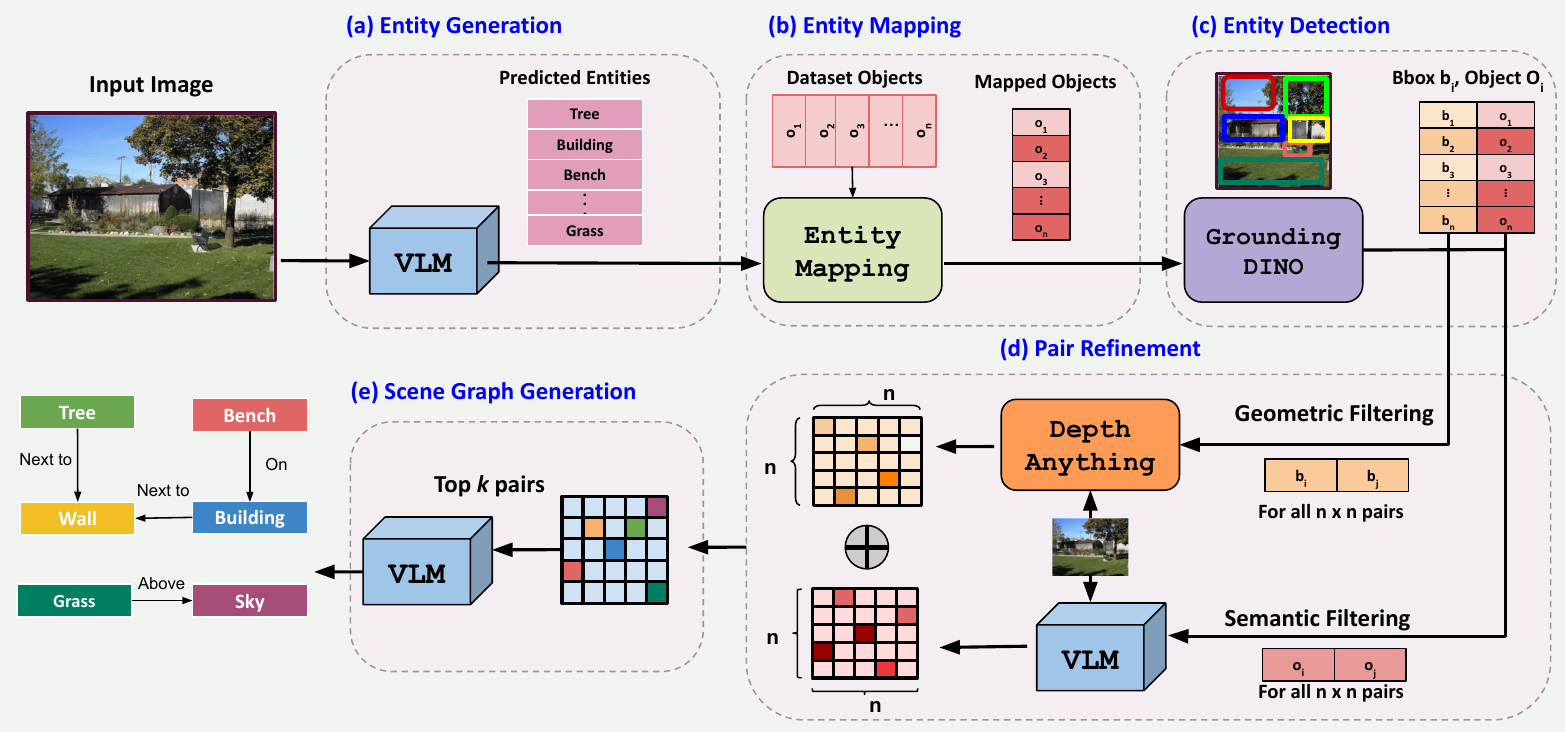}
  \caption{Overview of our proposed framework for open-world SGG using VLMs.}
  \label{fig:main_fig}
\end{figure*}

Figure \ref{fig:main_fig} shows our framework for Ow-SGG using VLMs, comprising of the following five steps.

% Our pipeline consists of four key stages: 
% \ak{\st{Don't need itemized list. Also can break it into 5 steps instead of 4 by describing Matching and Bounding Box Detection as two separate tasks. This overview paragraph should fully summarize Figure 2. We can move some text from later sections to here. Or we can remove this overview paragraph and directly get into the 4 or 5 steps. Should also mention that the prompts used for VLMs at every step are provided in the appendix.}}
% \vspace{-0.5em}
% \begin{enumerate}[left=0pt, labelwidth=0pt, align=left]
%     \item \textit{Entity Generation}: generates candidate objects or phrases from the input image $I$ using a vision-language model (VLM).
%     \vspace{-0.1em}
%     \item \textit{Entity Extraction and Detection}: filters out spurious candidates by performing embedding-based matching against a fixed set of dataset-defined object categories. The matched entities are then localized using Grounding DINO to obtain bounding boxes.
%     \vspace{-0.3em}
%     \item \textit{Pair Refinement}: reduces the combinatorial explosion of entity pairs using a hybrid filtering strategy based on semantic similarity and geometric proximity.
%     \vspace{-0.3em}
%     \item \textit{Scene Graph Generation}: employs the base VLM to generate relational triplets from the refined set of object pairs, forming the final scene graph.
% \end{enumerate}

\subsubsection{Entity Generation}
The goal of this step is to enumerate a diverse set of potential entities present in an image by prompting vision-language models (VLMs). We simply prompt a VLM to generate candidate object classes (or entities) present in an image.

% \ak{\st{Need to clarify terminologies: 1. entities, object labels, predicted entities, predicted labels mean the same thing. Should we consistently call it "detected entities". 2. GT labels, canonical object categories, dataset ontology, predefined vocabulary, predefined categories mean the same thing. Should we consistently call it "GT labels"?}}

\subsubsection{Entity Mapping}
In this step, we consider the task of mapping entities generated by VLMs to known object categories. Note that the entities predicted by a VLM may not directly correspond to the predefined or canonical class labels of objects in the dataset, and may occasionally include paraphrased or hallucinated concepts. Consequently, further processing is required to semantically align these free-form predictions of entity names with the predefined dataset object names. To achieve this, the entity mapping module uses an embedding-based matching mechanism to associate predicted entities to the dataset object categories. To identify semantically similar categories for every predicted entity, we compute similarity scores using a contrastive text encoder (SimCSE \cite{gao2021simcse}) and rank the candidates using a softmax-based scoring strategy. In practice, we retain up to $k$ dataset object categories whose similarity scores fall within a $\delta$-neighborhood of the top match with respect to any predicted entity. This strategy enhances robustness to synonyms or ambiguous naming (e.g., “man” vs. “boy”) while maintaining high semantic alignment. At the end of this process, every predicted entity is associated with one or more dataset object categories, enabling consistent downstream reasoning. The complete procedure is detailed in the \texttt{Semantic Pair Scoring for Entity Mapping} box below.

\subsubsection{Entity Detection}
Once the VLM-generated entities have been mapped to a list of candidate objects, Grounding DINO~\cite{liu2024grounding} is used to localize different instances of every object by predicting its corresponding bounding box inside the image. This step also serves as another layer of refinement by ignoring entities predicted but otherwise not present in the image. 

\subsubsection{Pair Refinement}
Given the set of object proposals \(\{(\mathbf{b_i}, o_i)\}_{i=1}^N\) produced by the Object Detection module, the next step in the pipeline involves constructing a list of meaningful object pairs \((\mathbf{v_i, v_j})\) for relational inference. 
% A na\"\i ve approach is to enumerate all possible object pairs. However, not all such pairs are semantically meaningful or spatially relevant. 
% To address this, we employ a pair refinement module that filters out noisy or irrelevant pairs, retaining only those likely to participate in valid relationships. However, the quality of this pair list significantly influences downstream relationship prediction.
A na\"\i ve approach is to select only overlapping or spatially adjacent objects. However, this has the risk of missing crucial object interactions that are semantically meaningful but show small geometric overlap (e.g., object pairs with relationships such as \texttt{looking at}). Conversely, enumerating all \(N(N-1)\) possible directed object pairs is computationally expensive and introduces a high degree of noise, especially when, 1) the resulting sequence becomes too long to be processed by  most VLMs, and 2) many object pairs are semantically or spatially irrelevant. \\
We propose a \textit{pair refinement module} that combines semantic and geometric cues to generate a concise yet informative set of candidate object pairs for relationship prediction before directly pruning out the object pairs. The module comprises two branches -- semantic pair refinement and geometric pair refinement -- which are then fused to produce a final refinement map.

\vspace{0.5em}
\noindent\textbf{Semantic pair refinement} utilizes a VLM to estimate the semantic closeness of interactions between every object pair. Given a pair \((o_i, o_j)\) of object category labels, the VLM assigns a semantic compatibility score:
\begin{equation}
\mathbf{P_{ij}^{S}} = \texttt{VLM}(o_i, o_j) \in [0, 1],
\end{equation}

where \(\mathbf{P_{ij}^{S}}\) is the semantic compatibility between objects \({i}\) and \({j}\). Since the VLM score is conditioned only on the object types (not their specific locations), we assign the same \(\mathbf{P_{ij}^S}\) to all instances of an object class pair \((o_i, o_j)\). This results in a semantic pair matrix \(\mathbf{P^{S}} \in \mathbb{R}^{N \times N}\).

\begin{tcolorbox}[title=\texttt{Semantic Pair Scoring for Entity Mapping}, colback=gray!5, colframe=black!50, fonttitle=\bfseries, top=1pt, bottom=1pt, before skip=4pt, after skip=4pt, boxsep=2pt, left=2pt]
Let $p$ be a predicted entity, $D = \{d_1, \dots, d_N\}$ the dataset object categories, $h(\cdot)$ a SimCSE encoder, $\tau=0.2$ the temperature, $\delta=0.05$ the threshold, and $k=2$ the maximum number of retained dataset object categories per entity.
\begin{enumerate}[left=1pt, itemsep=2pt]
\item Compute similarity scores:
% \[
% s_i = \cos\left(h(\texttt{"There is a }p\texttt{ in the image."}),\\, h(\texttt{"There is a }d_i\texttt{ in the image."})\right)
% \]
\vspace{-5pt}
\begin{align}
s_i = \cos\big(& h(\text{"There is a }p\text{ in the image."}), \notag \\
              & h(\text{"There is a }d_i\text{ in the image."}) \notag )
\end{align}

\item Apply temperature-scaled softmax: \\
\vspace{-15pt}
\begin{center}
 $
\hat{s}_i = \frac{\exp(s_i / \tau)}{\sum_{j=1}^N \exp(s_j / \tau)}
$   
\end{center}

\item Define near-maximum set:
\vspace{-5pt}
\begin{center}
$
S_{\max} = \max_i \hat{s}_i,\quad C = \left\{i : S_{\max} - \hat{s}_i < \delta \right\}
$ 
\end{center}

\item Return top-$k$ object labels from $C$: \\
\vspace{-15pt}
\begin{center}
$
M(p) = \{d_i \mid i \in C \text{ and } \hat{s}_i \text{ in top-}k \text{ of }C\}
$   
\end{center}

\item Combine mappings across all predicted entities: \\
\vspace{-15pt}
\begin{center}
$
\mathcal{M} = \bigcup_{p \in \mathcal{P}} M(p),
$ 
where \(\mathcal{P}\) is the set of predicted entities.
\end{center}

\end{enumerate}
The final output \(\mathcal{M}\) contains the set of matched dataset object categories aligned to the free-form predicted entities from the VLM.
\end{tcolorbox}

\vspace{0.5em}
\noindent\textbf{Geometric pair refinement} filters out spatially implausible pairs, by estimating spatial distances between objects using their 2D bounding boxes and an estimated depth map \(D\). Following \cite{prism}, 
% we project each 2D object's center into 3D space to obtain coordinates \(\mathbf{d}_i \in \mathbb{R}^3\).
% The geometric pair refinement module evaluates the spatial closeness of object pairs by combining 2D positional distance and estimated depth-based separation. For each object \(o_i\), 
we compute the 2D center of every bounding box \(\mathbf{b_i}\) as \(\mathbf{c}_i^{2D} = (x_i, y_i)\), and extract its normalized median depth value \(\mathbf{d_i} \in [0, 1]\) from a monocular depth map \(D\) that we generate using \cite{depthanything}.
Given two objects \(o_i\) and \(o_j\), we first compute the Euclidean distance between their 2D centers, $\mathbf{x_{ij}}$ $(= \| \mathbf{c}_i^{2D} - \mathbf{c}_j^{2D} \|_2)$,
and normalize it by the image diagonal length \(y = \sqrt{H^2 + W^2}\).
% , where \(H\) and \(W\) are the image height and width respectively.
The total distance between the objects is defined as a weighted combination of normalized 2D distance and absolute depth difference,
\begin{equation}
\mathbf{d_{ij}} = \lambda_1 \left(\frac{\mathbf{x_{ij}}}{y}\right) + \lambda_2  || \mathbf{d_i - d_j} ||_{2},
\label{eq:geom_scalar}
\end{equation}
where \(\lambda_1, \lambda_2 > 0\) are hyperparameters that control the relative contributions of the 2D and 3D components. A pair is retained if \(\mathbf{d_{ij}} < \tau\), for a chosen threshold \(\tau > 0\). To enable soft filtering, we convert this into a  compatibility score using a sigmoid function,
\begin{equation}
\mathbf{P_{ij}^{G}} = \sigma\left( -\beta (\mathbf{d_{ij}} - \tau) \right),
\label{eq:geom_gate}
\end{equation}
where \(\beta\) is an inverse-temperature parameter that controls the sharpness of the score, with higher values making the sigmoid more sensitive to whether the distance \(d_{ij} \) is above or below the threshold  \(\tau\). The resulting matrix \(\mathbf{P^G} \in \mathbb{R}^{N \times N}\) softly encodes spatial plausibility of each object pair.

\vspace{0.5em}
\noindent\textbf{Fusion of Semantic and Geometric Maps.}
Finally, we generate a unified refinement map \(\mathbf{P^{\text{final}}} \in \mathbb{R}^{N \times N}\) by taking a weighted sum of the semantic and geometric scores:
\begin{equation}
\mathbf{P_{ij}^{\text{combined}}} = \alpha \log \mathbf{P_{ij}^{S}} + (1 - \alpha) \log \mathbf{P_{ij}^{G}},
\end{equation}

where \(\alpha \in [0,1]\) is a tunable coefficient that controls the relative importance of semantic versus geometric refinement.
Top-$k$ pairs with the highest \(\mathbf{P_{ij}^{\text{combined}}}\) scores are retained for downstream relation prediction, resulting in a cleaner, more meaningful set of candidate triplets.

\subsubsection{Scene Graph Generation}
% After computing relationship scores for all candidate entity pairs using the pair refinement module, we select the top-\(k\) most promising pairs based on their relevance scores. 
We pass the set of refined pairs obtained from the previous module to a Vision-Language Model (VLM), which is prompted to predict the corresponding relationship between the two entities given the input image. This results in a set of relational triplets that form the final scene graph. 

\section{Experimental Results}
\label{sec:experiments}
\textbf{Evaluation Metrics.}  Having obtained the relationships $R$ and confidence scores $S$ for each object pair across the dataset, we measure performance against existing methods using two standard SG metrics: \textit{Recall@K (R@K)} and \textit{mean Recall@K (mR@K)}. These metrics quantify the proportion of ground-truth relationships a model is able to retrieve in its top-\(K\) predictions, either considering all predicate categories as a whole (\(R@K\)) or on a category-by-category basis (\(mR@K\)).
\\

% Recall@K computes the fraction of ground-truth relationships correctly predicted within the top-\(K\) ranked predictions. Given an image \( i \), let \( \mathcal{G}_i \) be the set of ground-truth relationships and \( \mathcal{P}_i^K \) the top-\(K\) predicted relationships. R@K is defined as:

% % \[
% %     \mathrm{R@K}_i = \frac{|\mathcal{G}_i \cap \mathcal{P}_i^K|}{|\mathcal{G}_i|} 
% % \]

% % Averaging over all \(N\) images:

% % \[
% %     \mathrm{R@K} = \frac{1}{N} \sum_{i=1}^{N} \mathrm{R@K}_i.
% % \]
% \[
% \mathrm{R@K}_i = \frac{|\mathcal{G}_i \cap \mathcal{P}_i^K|}{|\mathcal{G}_i|}, \quad \text{Averaging over all } N \text{ images: } \quad \mathrm{R@K} = \frac{1}{N} \sum_{i=1}^{N} \mathrm{R@K}_i
% \]
% meanRecall@K (mR@K) mitigates the bias toward frequent predicate categories by computing recall independently for each category. Let \( \mathcal{C}_i \) be the set of ground-truth predicate categories in image \( i \), and \( \mathcal{G}_{i,c} \) and \( \mathcal{P}_{i,c}^K \) represent the ground-truth and top-\(K\) predicted relationships for category \( c \), respectively. The per-category recall is:

% \[
% \mathrm{R@K}_{i, c} = \frac{|\mathcal{G}_{i,c} \cap \mathcal{P}_{i,c}^K|}{|\mathcal{G}_{i,c}|}, \quad \text{Averaging across all } N \text{ images: } \mathrm{mR@K} = \frac{1}{N} \sum_{i=1}^{N} \mathrm{mR@K}_i
% \]

\textbf{Datasets:} We perform extensive evaluations on the VG150 Dataset \cite{visualgenome}, Open Image v6 (OIV6) \cite{oiv6}, and the Panoptic Scene Graph Generation Dataset (PSG) \cite{psg}. The VG150 has 150 objects and 50 relationship categories. The OIV6 has 601 objects and 30 relationship categories while the PSG dataset has 133 object and 56 relationship categories. 
% \mridul{Add dataset linceses in supplymentary}
\\

\textbf{Backbone and Baselines:} We consider the following VLM backbones for implementing the proposed approach - LlaVa-next 7b \cite{llavanext}, Qwen2-vl 7b and Qwen2-vl 72b \cite{qwen2-vl}. We choose these VLMs as they are general foundational models and we evaluate them to exhibit the potential of our framework in being a model-agnostic SG Evaluation framework. We compare our evaluations against some well-known SOTA baselines such as - PGSG~\cite{from-pixels-to-graphs}, SGTR~\cite{sgtr}, RAHP~\cite{rahp}, OvSGTR~\cite{ovsgtr}.
\\

% \textbf{Tasks} \textcolor{red}{We consider the following tasks in this work,}
% \ak{This should be moved to Section 3 as it is also a contribution that will be  ignored as a known experiment setup if left here. For each OW setup, we should explain what previous works have explored it.}
% - CS (Closed-Set)
%   All triplets whose objects and predicates are within the training vocabulary. No filtering is applied.
  
% - ZS (Zero-Shot)
%   Triplets where the object–predicate–object combination has never appeared in the training set, although individual objects and predicates have been seen.

% - OVR (Open-Vocabulary Relations)
%   Triplets in which the predicate is novel (not seen during training). Objects may be seen or unseen.
  
% - OVD (Open-Vocabulary Object Detections)
%   Triplets in which the objects are novel (not seen during training). Predicates may be seen or unseen.

% - OVD+R (Open-Vocabulary Detection + Relations)
%   Union filter: triplets where predicate OR subject/object is novel. Allows unseen objects and predicates.

% - OW (Open World)
%    Triplets where both predicate AND subject/object are novel. Only triplets composed entirely of novel elements are included.OW $\subset$ OVDR: OW is a subset of OVDR (more restrictive on novelty).}

\subsection{Open Vocabulary Relationship (OVR) Results}
We evaluate our framework on the Open-Vocabulary Relationship Prediction (OVR) task, which assesses a model’s ability to correctly identify predicates that are absent from the training set. This task measures a model's capacity to generalize to rare or unseen relationships. Since our framework operates without any task-specific training, it is particularly well-suited for open-vocabulary settings. In contrast to conventional scene graph generation (SGG) models—which often struggle with unseen predicates due to their dependence on fixed label spaces—our approach leverages the semantic priors of vision-language models to reason over a broader predicate space. As shown in Table~\ref{tab:ovr_ovd}, our Qwen2-72B-based framework outperforms baseline methods on the OVR task for the PSG dataset. On the Visual Genome (VG) dataset under the PredCls setting, the OwSGG Qwen2-72B model achieves performance comparable to the best existing model. However, in most other settings—particularly on VG—our models underperform relative to baselines. These results suggest that vision-language models can generalize well in simpler but face challenges when applied to more complex or varied data.

\begin{table}[!htbp]
\caption{{\textbf{Close Vocabulary SGG Performance on VG150, OIV6, and PSG}: We show Zero-Shot and Close Vocabulary results on the VG150, OIV6 and the PSG Dataset. We compare our results on both SgDet and PredCls for VG150 and OIV6 and only SgDet for PSG.}}
% \tiny
\centering
\renewcommand{\arraystretch}{1.2}
\resizebox{\linewidth}{!}{
\begin{tabular}{l|l|l|lll}
\cmidrule{1-6}
\multicolumn{1}{c}{\multirow{2}{*}{\textbf{}}} &
  \multicolumn{1}{c}{\multirow{2}{*}{\textbf{}}} &
  \multicolumn{1}{c}{\multirow{2}{*}{\textbf{Method Name}}} &
  \multicolumn{2}{c}{\textbf{Close Vocabulary}} &
  \multicolumn{1}{c}{\textbf{Zero-Shot}} \\ \cmidrule{4-6} 
\multicolumn{1}{c}{} &
  \multicolumn{1}{c}{} &
  \multicolumn{1}{c}{} &
  \multicolumn{1}{c}{\textbf{mR @ 20 / 50 / 100}} &
  \multicolumn{1}{c}{\textbf{R @ 20 / 50 / 100}} &
  \multicolumn{1}{c}{\textbf{R @ 20 / 50 / 100}} \\ \cmidrule{1-6}
\multirow{16}{*}{\rotatebox[origin=c]{90}{VG}}  & \multirow{10}{*}{\rotatebox[origin=c]{90}{PredCls}} & IMP~\cite{xu2017scene}               & 11.7 / 14.8 / 16.1        & -- / 44.8 / 53.1       & \multicolumn{1}{c}{--} \\
&         & MOTIFS\cite{zellers2018neural}            & 11.7 / 14.8 / 16.1        & 58.5 / 65.2 / 67.1    & -- / 10.9 /14.5        \\
 % &
  %  &
  % \begin{tabular}[c]{@{}l@{}}TRANSFORMER +\\ HIERCOM\end{tabular} &
  % 17.9 / 26.6 / 32.2 &
  % 54.8 / 68.5 / 75.0 &
  % -- / 20.1 / 26.8 \\
    &         & VCTree+HIERCOM~\cite{heircom}    & 17.6 / 26.3 / 31.8        & 55.9 / 69.8 / 75.8    & -- / 17.8 / 24.8         \\
&         & CooK~\cite{cook}              & -- / 33.7 / 35.8       & -- / 62.1 / 64.2       & \multicolumn{1}{c}{--} \\
 &
   &
  CaCao~\cite{yu2023visually} &
  36.2 / 31.7 / 43.7 &
  \multicolumn{1}{c}{--} &
  \multicolumn{1}{c}{--} \\
    \cdashline{3-6}
    &         & OwSGG (llava-next) & \multicolumn{1}{c}{9.74 / 14.96 / 19.26}                     & \multicolumn{1}{c}{9.72 / 14.87 / 19.08}                     & \multicolumn{1}{c}{3.99 / 6.74 / 10.02}                     \\
    &         & OwSGG (Qwen7b)     & \multicolumn{1}{c}{4.82 / 8.73 / 12.64}                     & \multicolumn{1}{c}{ 4.9 / 8.9 /12.87}                     & \multicolumn{1}{c}{2.67 / 5.77 / 8.82}                     \\
    &         & OwSGG (Qwen72b)    & \multicolumn{1}{c}{7.63 / 13.54 / 19.76}                     & \multicolumn{1}{c}{7.53 / 13.44 / 19.64}                     & \multicolumn{1}{c}{3.3 / 6.52 / 9.7}                     \\ \cline{2-6} 
    & \multirow{6}{*}{\rotatebox[origin=c]{90}{SGDet}}   & SSRCNN~\cite{teng2022structured}            & -- / 18.6 / 22.5        & -- / 23.7 / 27.3        & -- / 3.1 / 4.5         \\
&         & SGTR~\cite{sgtr}       & -- / 12.0 / 15.2        & --/ 24.6 / 28.4        & -- / 2.5 / 5.8         \\
    &         & PGSG~\cite{from-pixels-to-graphs}              & -- / 8.9 / 11.5         & -- / 16.7 / 21.2        & -- / 6.2 / 8.5         \\
    \cdashline{3-6}
    &         & OwSGG (llava-next) & \multicolumn{1}{c}{1.88 / 2.89 / 3.7}                     & \multicolumn{1}{c}{1.7 / 2.61 / 3.36}                     & \multicolumn{1}{c}{0.98 / 1.71 / 2.36}                     \\
    &         & OwSGG (Qwen7b)     & \multicolumn{1}{c}{0.67 / 1.15 / 1.71}                     & \multicolumn{1}{c}{0.64 / 1.09 / 1.61}                     & \multicolumn{1}{c}{0.48 / 0.91 / 1.32}                     \\
    &         &    OwSGG (Qwen72b)    & \multicolumn{1}{c}{1.38/ 2.43 / 3.4}                     & \multicolumn{1}{c}{1.3 / 2.28 / 3.18}                     & \multicolumn{1}{c}{0.73 / 1.24 /1.95}                        \\ \hline
\multirow{12}{*}{\rotatebox[origin=c]{90}{OI6}} & \multirow{7}{*}{\rotatebox[origin=c]{90}{PredCls}} & SGTR~\cite{sgtr}              & \multicolumn{1}{c}{--} & -- / 59.9 / --          & \multicolumn{1}{c}{--} \\
    &         & ReIDN~\cite{reldn}             & \multicolumn{1}{c}{--} & -- / 72.8 / --          & \multicolumn{1}{c}{--} \\
    &         & GPS-Net~\cite{gpsnet}           & \multicolumn{1}{c}{--} & -- / 74.7 / --          & \multicolumn{1}{c}{--} \\
    &         & HEIRCOM~\cite{heircom}     & \multicolumn{1}{c}{--} & -- / 85.4 / --          & \multicolumn{1}{c}{--} \\
    \cdashline{3-6}
    &         & OwSGG (llava-next) & \multicolumn{1}{c}{59.92 / 66.82 / 70.24}                     & \multicolumn{1}{c}{59.88 / 66.81 / 70.22}                     & \multicolumn{1}{c}{30.08 / 35.03 / 36.59}                     \\
    &         & OwSGG (Qwen7b)     & \multicolumn{1}{c}{56.91 / 67.59 / 73.51}                     & \multicolumn{1}{c}{56.88 / 67.6 / 73.47}                     & \multicolumn{1}{c}{26.82 / 33.46 / 36.59}                     \\
    &         & OwSGG (Qwen72b)    & \multicolumn{1}{c}{71.54 / 79.83 / 83.76}                     & \multicolumn{1}{c}{71.56 / 79.86 / 83.78}                     & \multicolumn{1}{c}{40.1 / 47.14 / 47.14}                     \\ \cline{2-6} 
    & \multirow{5}{*}{\rotatebox[origin=c]{90}{SGDet}}   & SGTR~\cite{sgtr}              & -- /38.6 / -- & -- / 59.1 / --  & -- / 19.4 / 31.6       \\
    &         & PGSG~\cite{from-pixels-to-graphs}              & -- / 8.9 / 11.5 & -- / 16.7 / 21.2 & -- / 23.1 / 38.6       \\
    \cdashline{3-6}
    &         & OwSGG (llava-next) & \multicolumn{1}{c}{7.93 / 9.6 / 11.21}                     & \multicolumn{1}{c}{7.9 / 9.55 / 11.16}                     & \multicolumn{1}{c}{2.34 / 4.04 / 6.77}                     \\
    &         & OwSGG (Qwen7b)     & \multicolumn{1}{c}{2.7 / 4.61 / 6.74}                     & \multicolumn{1}{c}{2.68 / 4.59 / 6.71}                     & \multicolumn{1}{c}{2.47 / 2.99 / 2.99}                     \\
    &         & OwSGG (Qwen72b)    & \multicolumn{1}{c}{6.68 / 8.8 / 10.82}                     & \multicolumn{1}{c}{6.65 / 8.75 / 10.75}                     & \multicolumn{1}{c}{3.52 / 3.52 / 4.3}                     \\ \hline
\multirow{6}{*}{\rotatebox[origin=c]{90}{PSG}} & \multirow{6}{*}{\rotatebox[origin=c]{90}{SGDet}}   & PSGTR~\cite{psg}             & -- / 20.3 / 21.5 & -- / 32.1 / 35.3 & -- / 3.1 / 6.4         \\
    &         & SGTR~\cite{sgtr}      & -- / 24.3 / 27.2  & -- / 33.1 / 36.3 & -- / 4.1 / 5.8         \\
    &         & PGSG~\cite{from-pixels-to-graphs}           & -- / 20.9 / 22.1 & -- / 32.7 / 33.4 & -- / 6.8 / 8.9         \\
    \cdashline{3-6}
    &         & OwSGG (llava-next) & \multicolumn{1}{c}{5.59 / 8.12 / 10.02}                     & \multicolumn{1}{c}{5.63 / 8.12 / 10.08}                     & \multicolumn{1}{c}{1.84 / 2.51 / 4.68}                     \\
    &         & OwSGG (Qwen7b)     & \multicolumn{1}{c}{3.71 / 6.13 / 8.47}                     & \multicolumn{1}{c}{3.93 / 6.34 / 8.59}                     & \multicolumn{1}{c}{1.84 / 3.34 / 5.3}                     \\
    &         & OwSGG (Qwen72b)    & \multicolumn{1}{c}{7.22 / 10.49 / 13.67}                     & \multicolumn{1}{c}{7.36 / 10.68 / 13.98}                     & \multicolumn{1}{c}{2.84 / 4.68 / 6.44}                     \\ \hline
\end{tabular}
}
\label{tab:close_vocabulary}
\end{table}

\begin{table}[!htp]
\caption{{\textbf{Open Vocabulary Relation SGG Performance on VG150 and PSG}: We show OVR results on the VG150 and the PSG Dataset. We compare our results on both SgDet \& PredCls.}}
\renewcommand{\arraystretch}{1.2}
\resizebox{\linewidth}{!}{
\begin{tabular}{llllll}
\hline
                     &                                         & \multicolumn{2}{c}{\textbf{VG}}                & \multicolumn{2}{c}{\textbf{PSG}}               \\ \hline
\textbf{}        & \textbf{Method Name}                    & \multicolumn{2}{c}{\textbf{novel (Relation)}}  & \multicolumn{2}{c}{\textbf{novel (Relation)}}  \\ \cline{3-6} 
                     &                                         & \textbf{mR @ 50 / 100} & \textbf{R @ 50 / 100} & \textbf{mR @ 50 / 100} & \textbf{R @ 50 / 100} \\ \hline
\multirow{7}{*}{\rotatebox[origin=c]{90}{SGDet}}               
                     & \multicolumn{1}{l|}{VS3+RAHP~\cite{rahp}}           & \multicolumn{1}{c}{--}  &  3.75 / 5.12      & \multicolumn{1}{c}{--}  & \multicolumn{1}{c}{--}  \\
                     & \multicolumn{1}{l|}{OvSGTR~\cite{ovsgtr}}             & \multicolumn{1}{c}{1.82 / 2.32}  & 13.45 / 16.19     & \multicolumn{1}{c}{--} & \multicolumn{1}{c}{--} \\
                     & \multicolumn{1}{l|}{OvSGTR+RAHP~\cite{ovsgtr}}        & \multicolumn{1}{c}{3.01 / 4.04}  & 15.59 / 19.92   & \multicolumn{1}{c}{--}   & \multicolumn{1}{c}{--}    \\
                     & \multicolumn{1}{l|}{PGSG~\cite{from-pixels-to-graphs}}               &   \multicolumn{1}{c}{3.7 / 5.2}   & \multicolumn{1}{c}{--}  &  \multicolumn{1}{c}{7.4 / 11.3}  &  \multicolumn{1}{c}{--} \\
                     & \multicolumn{1}{l|}{SGTR~\cite{sgtr}}               &   \multicolumn{1}{c}{0.0 / 0.0}   & \multicolumn{1}{c}{--}  &  \multicolumn{1}{c}{0.0 / 0.0}  &  \multicolumn{1}{c}{0.0 / 0.0} \\
\cdashline{2-6}
                     & \multicolumn{1}{l|}{OwSGG (LLaVA-next)} & \multicolumn{1}{c}{2.34 / 3.04} & \multicolumn{1}{c}{2.33 / 3.04} & \multicolumn{1}{c}{8.27 / 10.4} & \multicolumn{1}{c}{8.31 / 10.49} \\
                     & \multicolumn{1}{l|}{OwSGG (Qwen7b)}     & \multicolumn{1}{c}{1.14 / 1.67} & \multicolumn{1}{c}{1.15 / 1.67} & \multicolumn{1}{c}{5.77 / 7.51} & \multicolumn{1}{c}{5.93 / 7.6} \\
                     & \multicolumn{1}{l|}{OwSGG (Qwen72b)}    & \multicolumn{1}{c}{2.19 / 3.07} & \multicolumn{1}{c}{2.19 / 3.06} & \multicolumn{1}{c}{10.25 / 13.35} & \multicolumn{1}{c}{10.42 / 13.54} \\
\cmidrule{1-6}
\multirow{7}{*}{\rotatebox[origin=c]{90}{PredCls}}               
                     & \multicolumn{1}{l|}{CaCao~\cite{yu2023visually}}     & \multicolumn{1}{c}{--} & \multicolumn{1}{c}{7.4 / 9.7} & \multicolumn{1}{c}{--} & \multicolumn{1}{c}{--} \\
                     & \multicolumn{1}{l|}{PGSG~\cite{from-pixels-to-graphs}} & \multicolumn{1}{c}{5.2 / 7.7} & \multicolumn{1}{c}{--} & \multicolumn{1}{c}{--} & \multicolumn{1}{c}{--} \\
                     & \multicolumn{1}{l|}{SGTR+RAHP~\cite{rahp}}          & \multicolumn{1}{c}{11.82 / 15.46} & \multicolumn{1}{c}{15.46 / 20.37} & \multicolumn{1}{c}{--} & \multicolumn{1}{c}{--} \\
\cdashline{2-6}
                     & \multicolumn{1}{l|}{OwSGG (LLaVA-next)} & \multicolumn{1}{c}{0.75 / 1.36 / 1.5} & \multicolumn{1}{c}{0.74 / 1.36 / 1.5} & \multicolumn{1}{c}{4.82 / 5.77} & \multicolumn{1}{c}{4.89 / 5.81} \\
                     & \multicolumn{1}{l|}{OwSGG (Qwen7b)}     & \multicolumn{1}{c}{0.44 / 1.2 / 2.12} & \multicolumn{1}{c}{0.44 / 1.19 / 2.11} & \multicolumn{1}{c}{4.02 / 5.29} & \multicolumn{1}{c}{4.03 / 5.31} \\
                     & \multicolumn{1}{l|}{OwSGG (Qwen72b)}    & \multicolumn{1}{c}{7.64 / 11.04} & \multicolumn{1}{c}{7.62 / 11.02} & \multicolumn{1}{c}{6.36 / 7.68} & \multicolumn{1}{c}{6.34 / 7.69} \\
\end{tabular}}
\label{tab:ovr_ovd}
\end{table}

%%%%%%%%%%%%%%%% TaBLE 3 %%%%%%%%%%%%%%         

\begin{table}[!htbp]
\caption{{\textbf{Open Vocabulary Detection and Open World SGG Performance on VG150}: We show results for the SgDet task on the VG150 Dataset. † indicates that the results were generated for this work.}}
\renewcommand{\arraystretch}{1.2}
\resizebox{\linewidth}{!}{
\begin{tabular}{lccc}
\hline
\textbf{Method Name}                    & \multicolumn{2}{c}{\textbf{OVD + R}}      & \textbf{OW}    \\ 
\cmidrule(lr){2-3} \cmidrule(lr){4-4}
                                        & \textbf{novel (Object)} & \textbf{novel (Relation)}   & \textbf{novel (Object \& Relation)} \\ 
                                        & \textbf{R@50 / R@100}   & \textbf{R@50 / R@100}       & \textbf{R@50 / R@100} \\ 
\hline
IMP~\cite{xu2017scene}                                     & 0.00 / 0.00             & 0.00 / 0.00                 & 0.00 / 0.00\textsuperscript{†} \\
MOTIFS~\cite{zellers2018neural}                                  & 0.00 / 0.00             & 0.00 / 0.00                 & 0.00 / 0.00\textsuperscript{†} \\
VCTREE~\cite{tang2019learning}                                  & 0.00 / 0.00             & 0.00 / 0.00                 & 0.00 / 0.00\textsuperscript{†} \\
TDE~\cite{tang2020unbiased}                                     & 0.00 / 0.00             & 0.00 / 0.00                 & 0.00 / 0.00\textsuperscript{†} \\
VS3~\cite{zhang2023learning}                                     & 6.00 / 7.51             & 0.00 / 0.00                 & --                     \\
OvSGTR (Swin-B)~\cite{ovsgtr}                         & 17.58 / 21.72           & 14.56 / 18.20               & 5.97 / 10.06\textsuperscript{†} \\
VS3+RAHP~\cite{rahp}                                & 13.01 / 14.82           & 3.75 / 5.12                 & --                     \\
OvSGTR+RAHP (Swin-T)~\cite{rahp}                    & 12.45 / 15.38           & 13.31 / 16.46               & --                     \\
\hline
OwSGG (LLaVA-next)                      & 2.37 / 3.07               & 2.33 / 3.04                 & 1.92 / 2.56            \\
OwSGG (Qwen7b)                          & 0.87 / 1.28             & 1.15 / 1.67                 & 0.86 / 1.18            \\
OwSGG (Qwen72b)                         & 1.88 / 2.73             & 2.19 / 3.06                 & 1.61 / 2.41            \\
\hline
\end{tabular}}
\label{tab:ovdr_ow}
\end{table}

\subsection{Close Vocabulary and Zero-Shot Results}
We also evaluate our open-world framework in the standard closed-vocabulary scene graph generation (SGG) setting, and additionally report zero-shot performance within this setup, as shown in Table~\ref{tab:close_vocabulary}. While our models are not expected to outperform baselines trained explicitly on dataset-specific labels, they yield several notable results. On the OIV6 dataset under the PredCls setting, the OwSGG (Qwen2-72B) model achieves higher recall at R@50 than all baselines except HEIRCOM~\cite{heircom}. As anticipated, our framework shows a relative advantage in the zero-shot scenario, where conventional models often fail to generalize beyond their training vocabulary. This is evident in the SgDet setting, where OwSGG (Qwen2-72B) surpasses all models except PGSG at R@100. However, on the more complex VG dataset, our models struggle to match baseline performance under both PredCls and SgDet settings.

\subsection{ Open Vocabulary Detection + Relation based SGG (OvD+R) and Open World Results}
The \textbf{OvD+R} setting evaluates models trained exclusively on base classes of objects and relationships, but tested on either novel objects or novel relationships—never both simultaneously. This setup measures partial generalization, where some components of the scene graph remain within the training distribution. In contrast, we introduce a more stringent \textbf{Open-World (OW)} evaluation setting, in which models must reason about both unseen objects and unseen relationships at test time, without any task-specific fine-tuning. The corresponding results are presented in Tab.~\ref{tab:ovdr_ow}. This scenario more closely reflects real-world conditions and provides a rigorous assessment of a model’s compositional generalization and robustness. Our proposed framework is explicitly designed for this open-world regime, leveraging the semantic flexibility of vision-language models without relying on supervision from predefined label sets. While the OwSGG results are still lower than those of baseline models trained with access to closed-world labels, they demonstrate the potential of this approach. The goal of establishing the OW baseline is to motivate future work toward models capable of operating effectively in fully unknown environments.

\begin{figure}[htbp]
\centering
\begin{minipage}[t]{0.47\textwidth}
    \includegraphics[width=\linewidth]{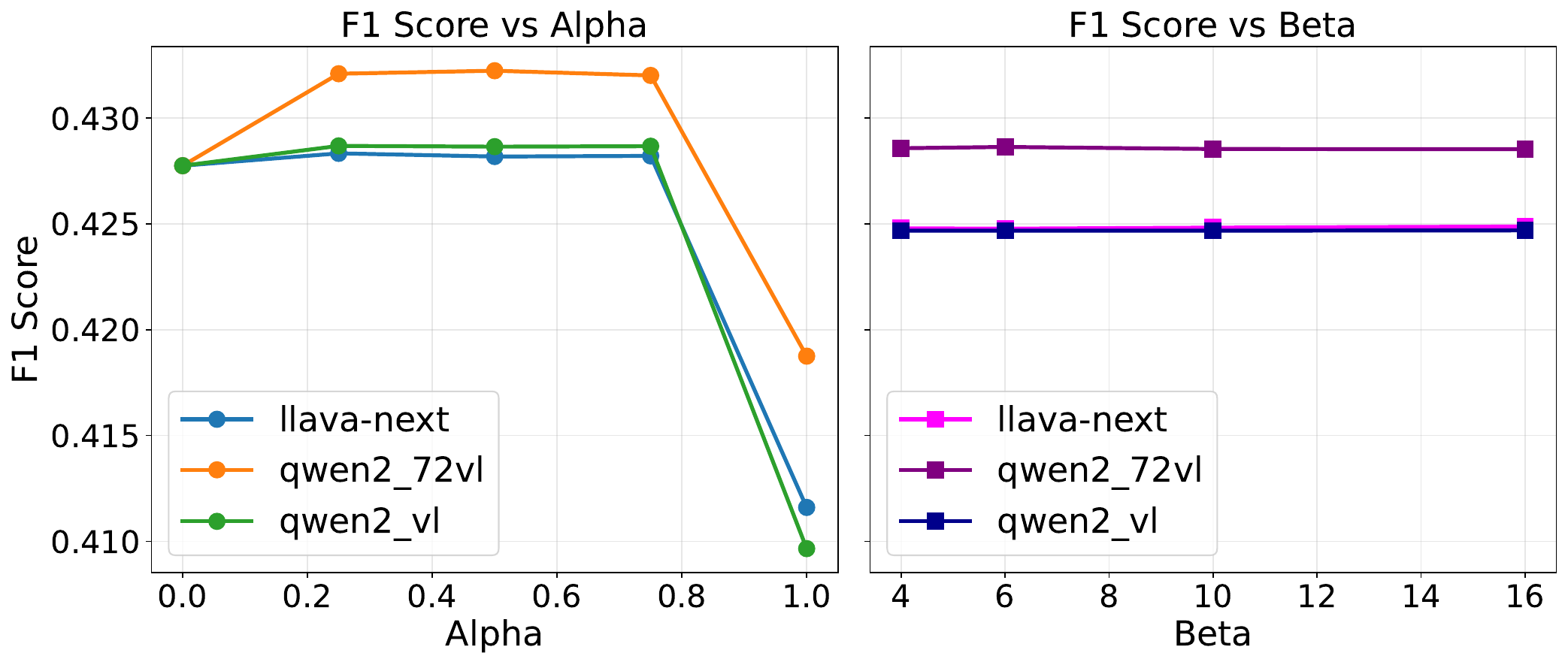}
    \captionof{figure}{
        \textcolor{black}{Ablation Study: F1 scores across different (a) $\alpha$ and (b) $\beta$ values for the Qwen-72B model.}
    }
    \label{fig:ablation_alpha_beta}
\end{minipage}
\hfill
\vspace{0.08cm}
\begin{minipage}[t]{0.48\textwidth}
    % \centering
    \small
    \captionof{table}{
        \textcolor{black}{Effect of depth and semantic filtering on PSG dataset (PredCls task, Qwen72b model). Bold = best, underline = $2^{nd}$ best.}
    }
    \label{tab:ablation}
    \resizebox{\linewidth}{!}{
    \begin{tabular}{cccccr}
        \cmidrule{1-5}
        % \toprule
        \textbf{Setup} & \textbf{Depth} & \textbf{Semantic} & \textbf{R@20/50/100} & \textbf{mR@20/50/100} & \\ \cmidrule{1-5}
        \multirow{3}{*}{CS}  & \checkmark & \xmark     & \textbf{4.8 / 6.94 / 8.32} & \textbf{4.76 / 6.86 / 8.22} & \\
                             & \xmark     & \checkmark & 2.67 / 4.03 / 5.09 & 2.61 / 4.0 / 5.05 & \\
                             & \checkmark & \checkmark & \ul{4.76 / 6.02 / 6.91} & \textbf{4.76}/ \ul{6.02 / 6.91} & \\ \cmidrule{1-5}
        \multirow{3}{*}{ZS}  & \checkmark & \xmark     & \ul{2.06} / \textbf{2.9} / \textbf{4.35}  & \ul{2.09} /\textbf{2.9} / \textbf{4.35}  & \\
                             & \xmark     & \checkmark & 1.34 / 2.17 / 2.17 & 1.34 / 2.17 / 2.17 & \\
                             & \checkmark & \checkmark & \textbf{2.34} / \ul{2.68} / \ul{3.51} & \textbf{2.34} / \ul{2.68} / \ul{3.51} & \\ \cmidrule{1-5}
        \multirow{3}{*}{OVR} & \checkmark & \xmark     & \ul{4.36 / 6.09 / 7.47} & \ul{4.4 / 6.17 / 7.55}  & \\
                             & \xmark     & \checkmark & 2.03 / 3.56 / 4.52 & 2.12 / 3.65 / 4.61 & \\
                             & \checkmark & \checkmark & \textbf{4.88 / 6.34 / 7.69} & \textbf{4.88 / 6.36 / 7.68} & \\
        % \bottomrule{1-5}
        \cmidrule{1-5}
    \end{tabular}}
\end{minipage}
\end{figure}

%%%%%%%%%%%%%% ABLATIONS %%%%%%%%%%%
\subsection{Ablation Results}

Fig.~\ref{fig:ablation_alpha_beta} (a) illustrates how varying the hyperparameter \(\alpha\), with a fixed \texttt{top\_k} of 25, influences model performance. As expected, increasing \texttt{top\_k} generally improves recall by increasing the likelihood of retrieving Ground Truth object pairs. The F1 score here reflects the quality of pair refinement—Recall captures how well Ground Truth pairs are preserved, while Precision indicates the degree to which noisy pairs are filtered out. An \(\alpha = 0\) corresponds to pure depth-based refinement, whereas \(\alpha = 1\) denotes purely semantic refinement. The results suggest that a balanced combination of both semantic and depth cues leads to more effective pair refinement for all the models. Fig.~\ref{fig:ablation_alpha_beta} (b) presents further ablation results, showing how the F1 score varies with different values of the \(\beta\) parameter used during depth-based refinement. These results highlight the model’s ability to retain meaningful pairs under a fixed \texttt{top\_k} of 25. Tab.~\ref{tab:ablation} complements these findings by reporting Triplet Recall values across various pair refinement strategies. We observe that while depth-only refinement performs well in a closed-vocabulary setting, combining semantic and depth-based filtering yields consistently better performance as the evaluation setting becomes more open and data-limited.

\section{Limitations and Conclusions}
\label{ch:conclusion}
The OwSGG framework leverages several pre-trained components—such as Grounding-DINO for object detection and SimCSE for embedding similarity—which can introduce error at various stages of the scene graph generation (SGG) pipeline. Understanding the contribution of these sources of error is an important direction for future work. Additionally, our method constructs textual prompts by pairing detected objects and feeding them into vision-language models (VLMs), which inherently limits scalability due to the context length constraints of these models. Future research can explore more efficient pair refinement strategies to reduce the number of candidate pairs that require evaluation by the VLM. Despite these limitations, our results demonstrate that VLMs, when guided by prompts and supplemented with object detection and embedding-based modules, are capable of predicting scene graph relationships without any task-specific training. This underscores the potential of zero-shot methods for structured vision-language tasks and paves the way toward more general, flexible, and interpretable visual reasoning systems.

\section{Acknowledgements}
We thank the Advanced Research Computing Center at Virginia Tech for GPU resources and the Ohio Supercomputer Center for additional compute support. This work was performed by UT-Battelle, LLC under DOE contract DE-AC05-00OR22725; the U.S. Government retains a non‐exclusive, irrevocable worldwide license to reproduce or publish this manuscript for government purposes, and results will be made publicly available per the DOE Public Access Plan (https://www.energy.gov/doe-public-access-plan). Support was also provided by NSF grant OAC-2118240 to the HDR Imageomics Institute.
{
    \small
    \bibliographystyle{ieeenat_fullname}
    \bibliography{main}
}

\appendix
\clearpage
\setcounter{page}{1}
\maketitlesupplementary

\section{Dataset Descriptions and Evaluation Splits}
\textbf{Datasets} We evaluate our framework on two categories of SGG: \textit{SgDet} and \textit{PredCls}. We evaluate on three datasets: Visual Genome (VG)~\cite{visualgenome}, Open Images V6 (OIV6)~\cite{oiv6}, and Panoptic Scene Graph (PSG)~\cite{psg}, using the standard splits from prior work~\cite{xu2017scene,zellers2018neural,psg}. Since our method requires no training, we evaluate only on the test data. For VG~\cite{visualgenome}, we follow the cleaning protocol of~\cite{xu2017scene,zellers2018neural}, removing images with insufficient annotations. This yields 26,446 test images (from 32,422), covering 150 object and 50 relation classes. For OIV6~\cite{oiv6}, we use the test split with 5,322 images, 601 object classes, and 30 relations. For PSG~\cite{psg}, we evaluate on the validation split, which contains 1,000 images, 133 objects, and 56 relations. 

We also leverage publicly available scripts and ID lists for split generation and novelty definitions:

\begin{itemize}
  \item \textbf{Zero-Shot Triplets} are generated using the T-CAR repository’s notebook%
    \protect\footnote{\url{https://github.com/jkli1998/T-CAR/blob/main/zs_check.ipynb}},
    which filters unseen triplets from the combined val+test pool.
  \item \textbf{VG Novel Predicates} (VG150) come from the OvSGTR codebase%
\protect\footnote{\url{https://github.com/gpt4vision/OvSGTR/blob/018453e07cf04be416ac42d13e1bf27d1611678d/datasets/vg.py\#L37}},
    and the base predicate set follows~\cite{he2022towards}.
  \item \textbf{OIV6 Novel Objects} are defined in the Pix2Grp CVPR2024 script%
    \protect\footnote{\url{https://github.com/SHTUPLUS/Pix2Grp_CVPR2024/blob/main/lavis/datasets/datasets/oiv6_rel_detection.py}},
    and similarly for \textbf{PSG Novel Predicates}%
    \protect\footnote{\url{https://github.com/SHTUPLUS/Pix2Grp_CVPR2024/blob/main/lavis/datasets/datasets/psg_rel_detection.py}}.
  \item For VG and OIV6, we adopt the train/val/test splits from previous works~\cite{xu2017scene,zellers2018neural}.  
    For PSG, we follow the official code and splits distributed at%
    \protect\footnote{\url{https://github.com/franciszzj/OpenPSG}}.
\end{itemize}

\section{Implementation Details}

\subsection{Vision Language Models}
All VLMs used are instruction-tuned to interpret structured prompts better. For inference, we leverage the vLLM framework~\cite{vllm}, which enables efficient execution of large-scale language models through a paged attention mechanism. Unlike traditional approaches that allocate contiguous memory, paged attention uses fixed-size pages, reducing fragmentation and improving memory reuse—allowing larger models to run with lower overhead. vLLM also features an optimized key-value (KV) cache that eliminates redundant computations by reusing previously computed attention values, significantly accelerating autoregressive generation. These optimizations make vLLM highly scalable and well-suited for low-latency inference with large VLMs. Due to hardware constraints, we quantize all models: 7B models from \texttt{float32} to \texttt{bfloat16}, and Qwen2-vl-72B using \texttt{AWQ}. This substantially reduces memory usage while maintaining performance.

\subsection{Entity Generation}
In the Entity Generation module, we prompt a VLM with the task of generating a comprehensive list of entities present in the input image. The module is configured using the following hyper-parameters:
\begin{enumerate}[itemsep=2pt, leftmargin=10pt]
    \item \texttt{num\_outputs=1}: We request a single generation output per image.
    \item \texttt{temperature=0.1}: A low temperature ensures deterministic outputs, reducing randomness and encouraging factual extraction.
    \item \texttt{max\_tokens=512}
    \item \texttt{top\_p=1.0}: This enables nucleus sampling with a large cutoff to avoid premature truncation of less frequent but relevant entities.
    \item \texttt{presence\_penalty=0.4}: Penalizes repetitions to encourage novel mentions without being too aggressive.
    \item \texttt{repetition\_penalty=1.1}: Mildly discourages duplicate tokens during generation.
\end{enumerate}

\textbf{Prompt examples.} We use dataset-specific prompts tailored to encourage comprehensive object enumeration. The prompt examples used for three datasets are shown by \texttt{PSG Dataset Prompt, Open Images (OI) Prompt} and \texttt{Visual Genome (VG) Prompt}.

\begin{tcolorbox}[title=\texttt{PSG Dataset Prompt}, colback=gray!5, colframe=black!50, fonttitle=\bfseries, top=1pt, bottom=1pt, before skip=4pt, after skip=4pt, boxsep=2pt, left=2pt]

\small\ttfamily

\#\#\# Task Start
You are an expert at detecting objects in images. You are given an image. Your task is to list all objects visible in the image, including both foreground and background. The objects may include natural elements, human-made structures, or any other discernible entities.

\#\#\# Output Format Instructions

- Do not repeat object names.\\
- Do not describe attributes, adjectives, or relationships. \\
- Return the result as a comma-separated list. \\
- If unsure, include it. \\
\#\#\# Prompt

List all the objects visible in the image, including foreground and background. Return the objects as a comma-separated list.

\end{tcolorbox}

\begin{tcolorbox}[title=\texttt{Open Images (OI) Prompt}, colback=gray!5, colframe=black!50, fonttitle=\bfseries, top=1pt, bottom=1pt, before skip=4pt, after skip=4pt, boxsep=2pt, left=2pt]
\small\ttfamily
\#\#\# Task Start

You are an expert at detecting objects in images. You are given an image. Your task is to identify and list all visible objects in the image, including both foreground and background. Include a wide range of recognizable categories, whether specific or general, as long as they are visibly present in the scene.

\#\#\# Output Format Instructions

- Do not repeat object names.  \\
- Do not describe attributes, adjectives, or relationships. \\  
- Return the result as a comma-separated list.  \\
- If unsure, include it.

\#\#\# Prompt

List all the objects visible in the image, including foreground and background. Return the objects as a comma-separated list.
\end{tcolorbox}

\begin{tcolorbox}[title=\texttt{Visual Genome (VG) Prompt}, colback=gray!5, colframe=black!50, fonttitle=\bfseries, top=1pt, bottom=1pt, before skip=4pt, after skip=4pt, boxsep=2pt, left=2pt]
\small\ttfamily
\#\#\# Task Start

You are an expert at detecting objects in images. You are given an image. Your task is to list all identifiable objects visible in the image, including those in the foreground and background. Include both whole objects and meaningful parts or components that are visually discernible.

\#\#\# Output Format Instructions

- Do not repeat object names.  \\
- Do not describe attributes, adjectives, or relationships. \\  
- Return the result as a comma-separated list.  \\
- If unsure, include it.

\#\#\# Prompt

List all the objects visible in the image, including foreground and background. Return the objects as a comma-separated list.
\end{tcolorbox}

% \newpage

\subsection{Entity Mapping}
\label{sec:ent_mapping}

Our entity‐mapping pipeline aligns VLM‐predicted object labels to a fixed ground‐truth vocabulary via a three‐stage cascade. First, each label is normalized (converted to lowercase, trimmed of whitespace, and stripped of all punctuation). Second, we compare the normalized prediction directly against a cache of normalized ground‐truth entries; any exact hits are accepted with confidence 1.0. Third, any remaining labels are resolved via semantic matching with a contrastively pretrained SimCSE \cite{gao2021simcse} encoder.

In the semantic stage, we convert each candidate label \emph{X} into a full sentence of the form
\begin{quote}
  “There is a \emph{X} in the image.”
\end{quote}
and embed it with SimCSE. We compare that embedding—via cosine similarity—to a cache of precomputed embeddings for every normalized ground‐truth entry. To sharpen the score distribution, we apply temperature scaling with \(\tau = 0.2\). We then filter out any ground‐truth entries whose cosine score falls more than \(\Delta = 0.05\) below the maximum observed score, and finally select the top \(k = 2\) remaining candidates as our matches.

\paragraph{Illustrative Mapping Cases}

We present examples to illustrate both positive and negative mapping outcomes from our entity alignment module. A mapping is considered \textbf{positive} if one or more of the matched categories appear in the ground truth, and \textbf{negative} if all matches are semantically reasonable but absent from the GT labels.

\vspace{0.5em}
\noindent\textbf{Positive Mapping Cases}
\begin{itemize}
  \item \emph{GT objects:} \texttt{person}, \texttt{tree}, \texttt{car}
  \item \emph{VLM prediction:} \texttt{man}
  \item \emph{SimCSE top-2 matches:}
    \begin{itemize}
      \item \texttt{gentleman} (cos = 0.92) \quad [not in GT] \\
      \item \texttt{person} (cos = 0.89) \quad [\textbf{in GT}]
    \end{itemize}

  \item \emph{VLM prediction:} \texttt{woman}
  \item \emph{SimCSE top-2 matches:}
    \begin{itemize}
      \item \texttt{lady} (cos = 0.90) \quad [not in GT] \\
      \item \texttt{person} (cos = 0.87) \quad [\textbf{in GT}]
    \end{itemize}

  \item \emph{GT objects:} \texttt{dog}, \texttt{grass}
  \item \emph{VLM prediction:} \texttt{puppy}
  \item \emph{SimCSE top-2 matches:}
    \begin{itemize}
      \item \texttt{canine} (cos = 0.82) \quad [not in GT] \\
      \item \texttt{dog} (cos = 0.79) \quad [\textbf{in GT}]
    \end{itemize}
\end{itemize}

\vspace{0.5em}
\noindent\textbf{Negative Mapping Cases}
\begin{itemize}
  \item \emph{GT objects:} \texttt{person}, \texttt{car}, \texttt{tree}
  \item \emph{VLM prediction:} \texttt{skateboarder}
  \item \emph{SimCSE top-2 matches:}
    \begin{itemize}
      \item \texttt{skateboard} (cos = 0.76) \quad [not in GT] \\
      \item \texttt{rider} (cos = 0.73) \quad [not in GT]
    \end{itemize}

  \item \emph{GT objects:} \texttt{tennis racket}, \texttt{person}
  \item \emph{VLM prediction:} \texttt{tennis player}
  \item \emph{SimCSE top-2 matches:}
    \begin{itemize}
      \item \texttt{athlete} (cos = 0.81) \quad [not in GT] \\
      \item \texttt{player} (cos = 0.78) \quad [not in GT]
    \end{itemize}
\end{itemize}

In Sec.~\ref{sec:detection} we show how the negative mapping cases are handled by using Grounding DINO~\cite{liu2024grounding} as our object detection module.

\subsubsection{Entity Mapping Ablation}
\begin{figure*}[htbp]
  \centering
  \includegraphics[width=0.75\textwidth]{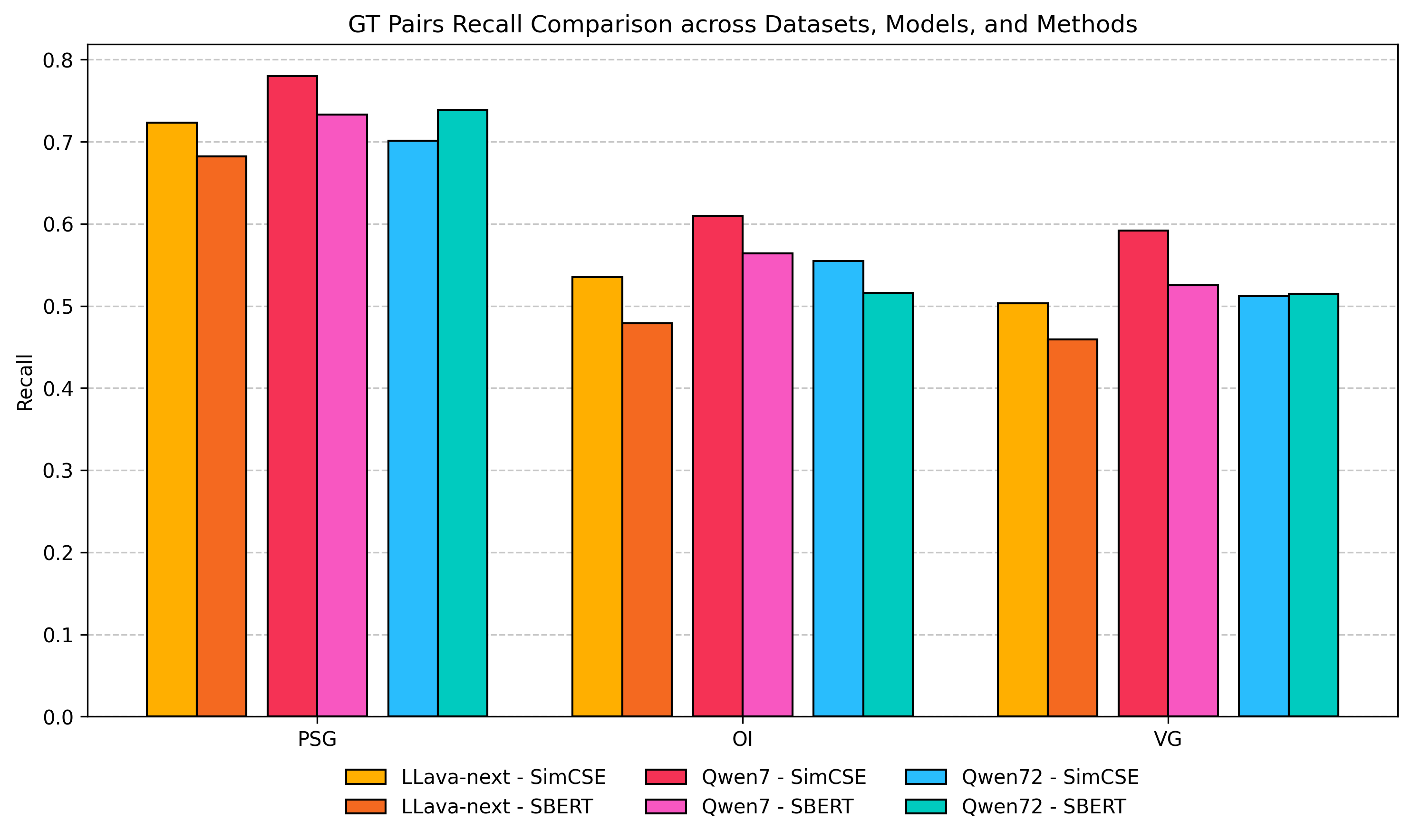}
  \caption{Recall comparison across datasets, models, and methods.}
  \label{fig:recall-comparison}
\end{figure*}

To quantify the benefit of SimCSE’s contrastive training, we ran an ablation comparing it against a standard Sentence-BERT (SBERT) \cite{reimers2019sentence} encoder—while keeping the same normalization and synonym steps across three datasets (PSG, OI, VG) and three VLMs: LLava Next, Qwen2-vl 7b \textit{Qwen7)} and Qwen2-vl 72b\text{(Qwen72)}. The grouped bar chart above shows recall for each model–method pairing. Overall, SimCSE (gold, crimson, sky-blue bars) yields up to a 5\% recall boost over SBERT (orange, pink, teal bars) on the PSG and OI sets, particularly for Qwen7, highlighting its stronger discrimination of fine-grained object labels. On the more challenging VG data, both methods converge to lower recall, although SBERT slightly outperforms SimCSE for Qwen72 on PSG. These results suggest that contrastive supervision in SimCSE enhances generalization in complex scenes, while SBERT can sometimes better capture subtle category nuances in smaller models as shown in Fig.~\ref{fig:recall-comparison}.

\subsection{Entity Detection}
\label{sec:detection}
\begin{figure}[htbp]
  \centering
  \includegraphics[width=0.45\textwidth]{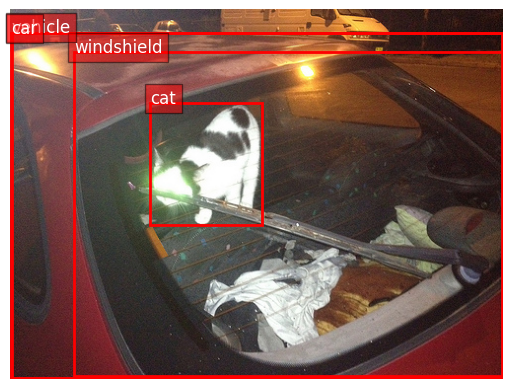}
  \caption{\textit{Example 1} — Initial mapped entities: \texttt{[windshield, vehicle, light, building, car, street, cat, bag]}. Irrelevant objects such as \texttt{light}, \texttt{building}, \texttt{street}, and \texttt{bag} are successfully filtered out.}
  \label{fig:filtering1}
\end{figure}

\begin{figure}[htbp]
  \centering
  \includegraphics[width=0.45\textwidth]{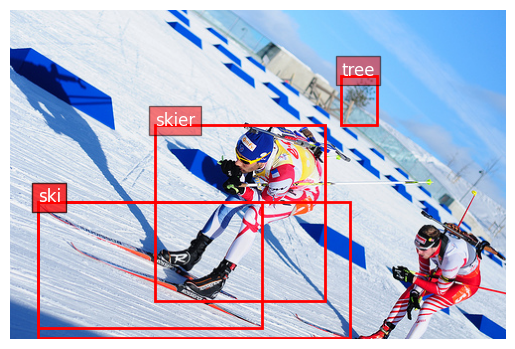}
  \caption{\textit{Example 2} — Initial mapped entities: \texttt{[ski, light, tree, skier, number, snow, roof]}. Irrelevant objects such as \texttt{light}, \texttt{number}, and \texttt{roof} are removed during filtering.}
  \label{fig:filtering2}
\end{figure}

We utilize Grounding-DINO \cite{liu2024grounding} for zero-shot entity detection, specifically employing the \emph{groundingdino\_swinb\_cogcoor} variant. We set the box\_threshold to $35\%$ and the text\_threshold to $25\%$, following default values recommended by the authors. A single object name serve as a single text prompt for Grounding-DINO.
\\
It is worth noting that the original Grounding-DINO paper highlights its capability to ground multiple objects in the text by separating their names with dots (e.g., `person.cat.dog'). However, in our practical experience, while combining multiple objects in a single prompt speeds up entity detection, it compromises the quality of detected boxes. Grounding-DINO demonstrates superior performance when tasked with detecting a single object per text prompt. Therefore, we adopt a strategy of providing individual object names to maximize detection quality. 
\\
\textbf{Object Filtering} As discussed in Section~\ref{sec:ent_mapping}, the entity mapping stage may generate spurious or semantically irrelevant object labels. Here, we show how the pair refinement and filtering stages effectively remove such cases before the final triplet prediction. Figures~\ref{fig:filtering1} and~\ref{fig:filtering2} illustrate two examples where several incorrect or irrelevant mapped entities are successfully discarded.

\newpage

\subsection{Pair Refinement}
We present the prompt formulation and hyperparameter values used in the two stages of pair refinement in our framework.

\subsubsection{Semantic Pair Refinement}
\begin{figure}[htbp]
  \centering
  \includegraphics[width=0.5\textwidth]{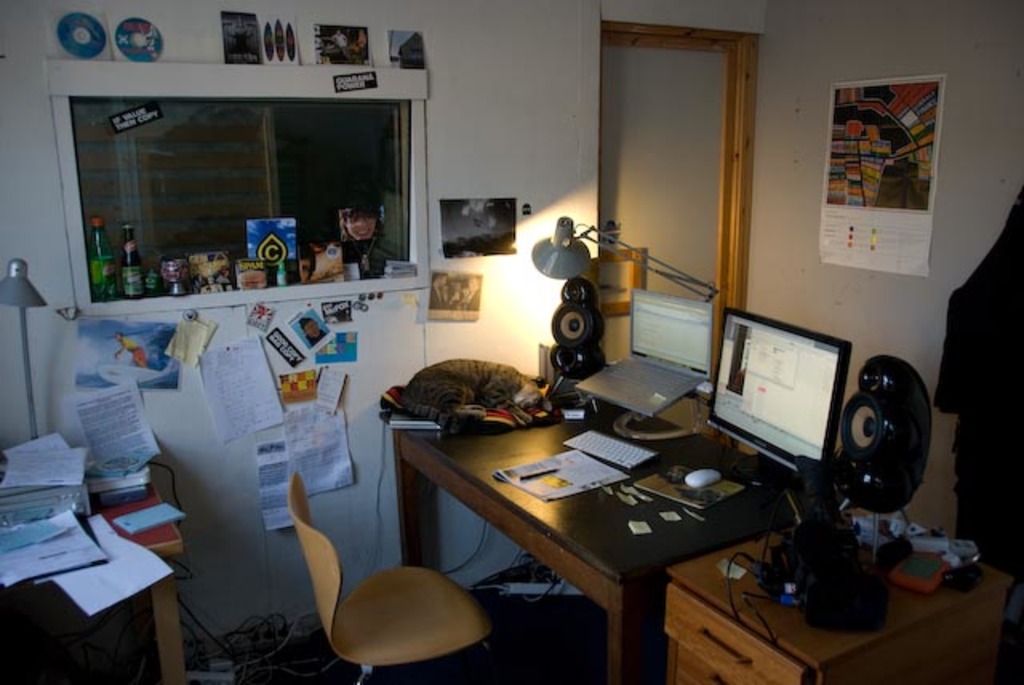}
  \caption{Example of semantic pair refinement. Given an image and a list of object pairs, the VLM is prompted to assign interaction likelihood scores, helping filter out semantically implausible relationships.}

  \label{fig:refinement}
\end{figure}

To perform semantic filtering, we present the VLM with the full set of candidate entity pairs and ask it to rank them by their semantic relevance. Fig.~\ref{fig:my-prompt} shows the exact prompt used in the example of Fig.~\ref{fig:refinement}, illustrating how the model refines the image’s relationships. 
% \textcolor{red}{Refer to the prompt and the image}

\subsubsection{Geometric Pair Refinement}

Following prior work~\cite{prism}, we adopt the same geometric distance formulation: \\
\(\lambda_1 \left(\frac{\mathbf{x_{ij}}}{y}\right) + \lambda_2 \| \mathbf{d_i - d_j} \|_2 < \tau\), 
where \(\lambda_1 = 1.0\), \(\lambda_2 = 1.5\), and \(\tau = 0.5\). Unlike~\cite{prism}, which directly prunes pairs exceeding this threshold, we convert the distance into a soft compatibility score using a sigmoid function (Eq.~\ref{eq:geom_gate}). 

We introduce an additional hyperparameter \(\beta\), which controls the sharpness of this score. We use \(\beta = 16\) for 7B models (LLaVA-next and Qwen2-vl 7B), and \(\beta = 10\) for Qwen2-vl 72B. For the final fusion of semantic and geometric scores, we set the weighting factor \(\alpha = 0.25\).

\subsection{Scene Graph Generation}
\begin{figure}[htbp]
  \centering
  \includegraphics[width=0.5\textwidth]{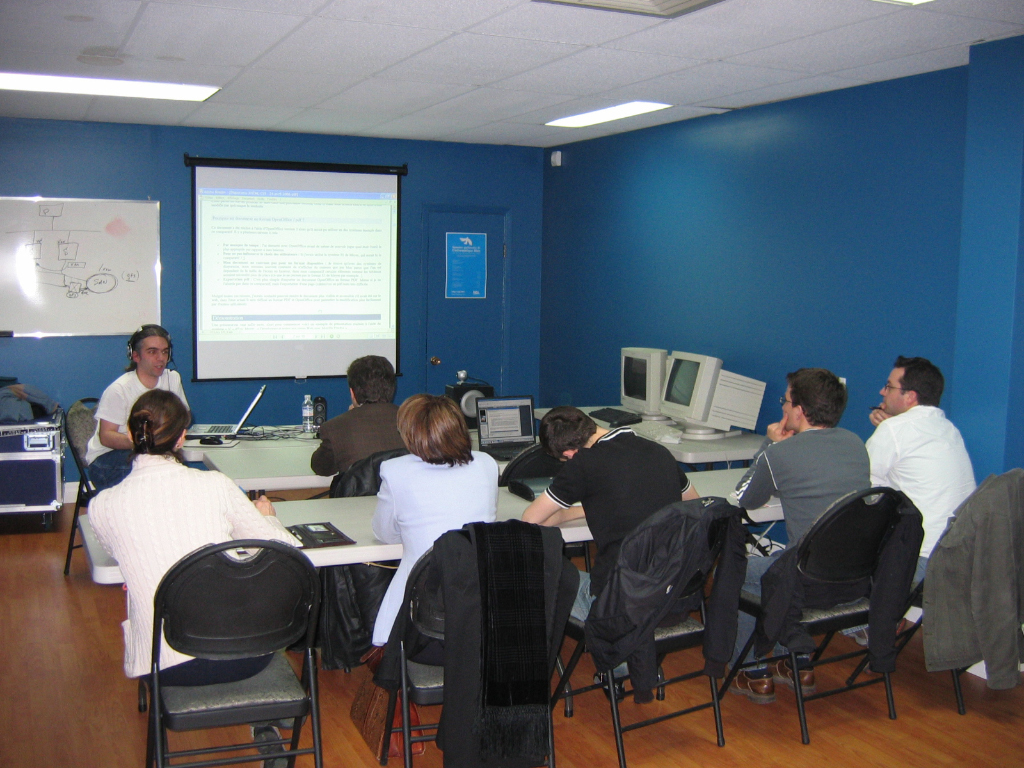}
  \caption{Final scene graph generation setup. Refined object pairs, along with their bounding box coordinates, are passed to the VLM to predict relationships.}

  \label{fig:sg_gen}
\end{figure}

In the final scene‐graph generation stage, we feed the VLM the semantically refined object pairs and ask it to infer their relationships. Because Qwen2-VL was instruction-tuned on bounding‐box annotations—unlike the LLaVA models—the precise prompt templates differ: see \texttt{Relation Generation Prompt for LLaVA} and \texttt{Relation Generation Prompt for Qwen2-VL} for the exact prompts used for Fig.~\ref{fig:sg_gen}. The final list of \textcolor{green}{correct} and \textcolor{red}{incorrect} outputs is presented in \texttt{Sample VLM Outputs (Correct and Incorrect)}. 

\newpage

\begin{tcolorbox}[title=Sample VLM Outputs (Correct and Incorrect), colback=gray!3, colframe=black!40, fonttitle=\bfseries, coltitle=black, left=2pt, right=2pt, boxsep=4pt]

\colorbox{green!20}{%
\parbox{\linewidth}{%
\textbf{Pair 1:} \\
Sentence1: The woman is sitting on the chair. | Sentence2: The chair is being used by the woman.}}

\vspace{4pt}
\colorbox{green!20}{%
\parbox{\linewidth}{%
\textbf{Pair 2:} \\
Sentence1: The woman is next to the chair. | Sentence2: The chair is beside the woman.}}

\vspace{4pt}
\colorbox{red!20}{%
\parbox{\linewidth}{%
\textbf{Pair 3:} \\
Sentence1: The woman is located on the table. | Sentence2: The table is behind the woman.}}

\vspace{4pt}
\colorbox{green!20}{%
\parbox{\linewidth}{%
\textbf{Pair 4:} \\
Sentence1: The woman is resting her arm on the table. | Sentence2: The table is supporting the woman’s arm.}}

\vspace{4pt}
\colorbox{red!20}{%
\parbox{\linewidth}{%
\textbf{Pair 5:} \\
Sentence1: The chair is on top of the table. | Sentence2: The table is on the chair.}}

\vspace{4pt}
\colorbox{green!20}{%
\parbox{\linewidth}{%
\textbf{Pair 6:} \\
Sentence1: The man is seated at the table. | Sentence2: The table is in front of the man.}}

\end{tcolorbox}

\section{Qualitative Results for Pair Refinement}

\begin{figure*}[t]
    \centering
    
    % The image is now on its own, centered, and can have its own width
    \includegraphics[width=0.5\textwidth]{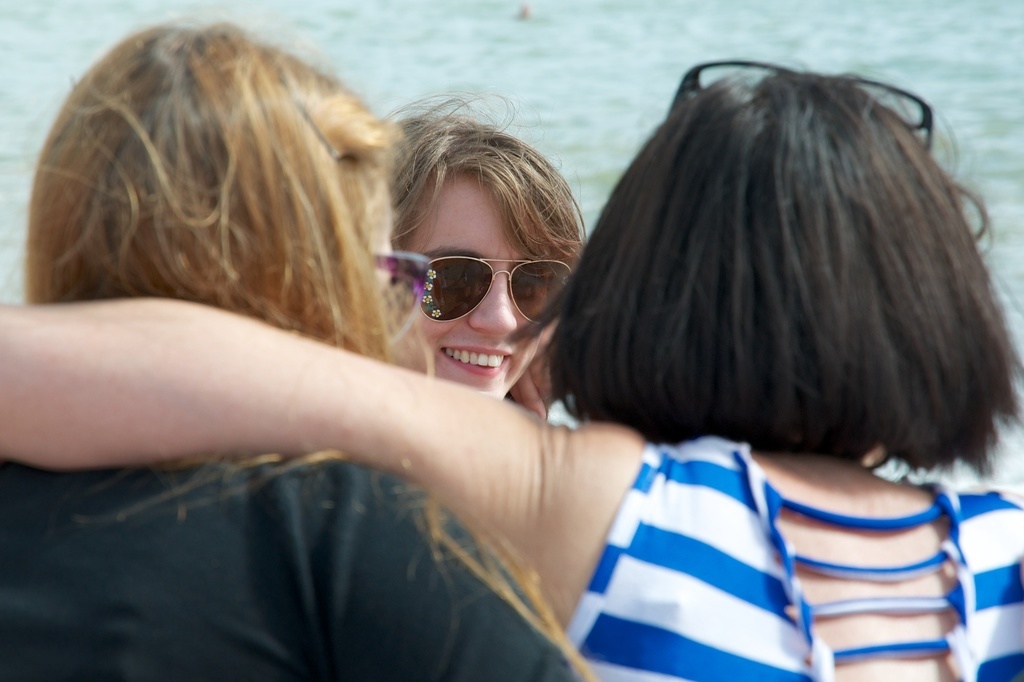}
    
    % Add some vertical space between the image and the table
    \vspace{1.5em} 
    
    % The table now uses the full available width
    \scriptsize\ttfamily
    \begin{tabular}{
        % This column definition automatically adapts to the full text width
        *{4}{L{\dimexpr.25\linewidth-2\tabcolsep}}
    }
        \toprule
        \textbf{Semantic} & \textbf{Depth} & \textbf{Fused} & \textbf{GT Pairs} \\
        \midrule

        % Row 1
        \cellcolor{red!20}\shortstack[l]{girl[385,79,587,399]\\glasses[660,57,936,146]} &
        \cellcolor{red!20}\shortstack[l]{sunglasses[416,256,572,324]\\goggles[413,254,575,325]} &
        \cellcolor{green!20}\shortstack[l]{girl[385,79,587,399]\\glasses[418,256,572,324]} &
        \cellcolor{green!20}\shortstack[l]{girl[385,79,587,399]\\glasses[418,256,572,324]} \\
\\
        % Row 2
        \cellcolor{green!20}\shortstack[l]{sunglasses[416,256,572,324]\\girl[385,79,587,399]} &
        \cellcolor{green!20}\shortstack[l]{glasses[418,256,572,325]\\girl[385,79,587,399]} &
        \cellcolor{green!20}\shortstack[l]{sunglasses[416,256,572,324]\\girl[385,79,587,399]} &
        \cellcolor{green!20}\shortstack[l]{sunglasses[416,256,572,324]\\girl[385,79,587,399]} \\
\\
        % Row 3
        \cellcolor{red!20}\shortstack[l]{girl[0,10,595,682]\\glasses[660,57,936,146]} &
        \cellcolor{green!20}\shortstack[l]{woman[380,90,587,417]\\girl[385,79,587,399]} &
        \cellcolor{green!20}\shortstack[l]{woman[380,90,587,417]\\girl[385,79,587,399]} &
        \cellcolor{green!20}\shortstack[l]{woman[380,90,587,417]\\girl[385,79,587,399]} \\

        \bottomrule
    \end{tabular}

    % The single caption now describes both the image above and the table
    \caption{Example 1: Qualitative comparison of top object pairs from different methods versus the Ground Truth (GT) pairs for the given image (top). The table (bottom) details these pairs. Green indicates a correct pair, while red indicates an incorrect one.}
    \label{fig:qualitative_refinement1}
\end{figure*}

\begin{figure*}[t]
    \centering
    
    % The image is placed first
    \includegraphics[width=0.5\textwidth]{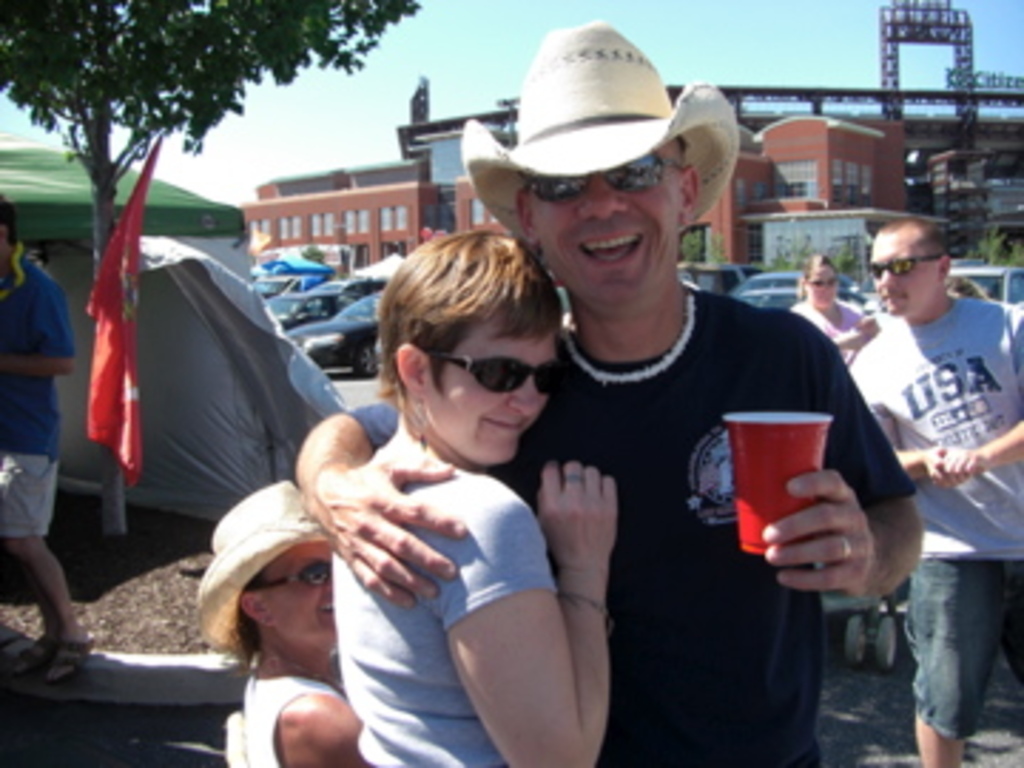}
    
    % Add vertical space for clean separation
    \vspace{1.5em} 
    
    % The table follows, using the full available width
    \scriptsize\ttfamily
    \begin{tabular}{
        % This column definition automatically adapts to the full text width
        *{4}{L{\dimexpr.25\linewidth-2\tabcolsep}}
    }
        \toprule
        \textbf{Semantic Pairs} & \textbf{Depth Pairs} & \textbf{Fused Pairs} & \textbf{GT Pairs} \\
        \midrule

        % Row 1
        \cellcolor{red!20}\shortstack[l]{girl[329,219,620,768]\\glasses[861,253,944,281]} &
        \cellcolor{green!20}\shortstack[l]{woman[329,219,620,765]\\girl[329,219,620,768]} &
        \cellcolor{green!20}\shortstack[l]{woman[329,219,620,765]\\girl[329,219,620,768]} &
        \cellcolor{green!20}\shortstack[l]{woman[329,219,620,765]\\girl[329,219,620,768]} \\
\\
        % Row 2
        \cellcolor{green!20}\shortstack[l]{girl[329,219,620,768]\\sunglasses[520,153,662,204]} &
        \cellcolor{green!20}\shortstack[l]{glasses[523,159,685,201]\\girl[329,219,620,768]} &
        \cellcolor{green!20}\shortstack[l]{sunglasses[520,153,662,204]\\girl[329,219,620,768]} &
        \cellcolor{green!20}\shortstack[l]{sunglasses[520,153,662,204]\\girl[329,219,620,768]} \\
\\
        % Row 3
        \cellcolor{red!20}\shortstack[l]{girl[329,219,620,768]\\sun hat[460,22,736,238]} &
        \cellcolor{green!20}\shortstack[l]{glasses[423,355,566,395]\\man[295,19,924,768]} &
        \cellcolor{green!20}\shortstack[l]{glasses[423,355,566,395]\\man[295,19,924,768]} &
        \cellcolor{green!20}\shortstack[l]{glasses[423,355,566,395]\\man[295,19,924,768]} \\

        \bottomrule
    \end{tabular}

    % A single, combined caption for both the image and the table
    \caption{Example 2: Qualitative comparison of top object pairs from different methods versus the Ground Truth (GT) pairs for the given image (top). The table (bottom) details these pairs. \textcolor{green!50!black}{Green} indicates a correct pair, while \textcolor{red}{red} indicates an incorrect one.}
    \label{fig:qualitative_refinement2}
\end{figure*}

\begin{figure*}[t]
    \centering
    
    % The image is placed first
    \includegraphics[width=0.5\textwidth]{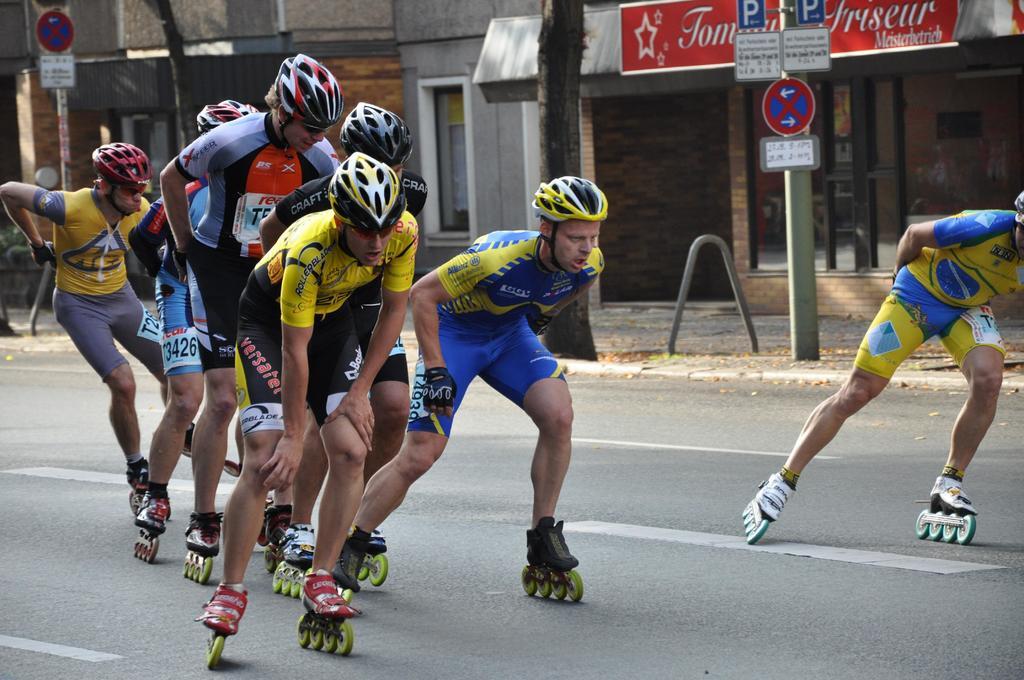}
    
    % Add vertical space for clean separation
    \vspace{1.5em} 
    
    % The table follows, using the full available width
    \scriptsize\ttfamily
    \begin{tabular}{
        % This column definition automatically adapts to the full text width
        *{4}{L{\dimexpr.25\linewidth-2\tabcolsep}}
    }
        \toprule
        \textbf{Semantic Pairs} & \textbf{Depth Pairs} & \textbf{Fused Pairs} & \textbf{GT Pairs} \\
        \midrule

        % Row 1
        \cellcolor{red!20}\shortstack[l]{bicycle helmet[90,138,156,218]\\man[751,189,1022,520]} &
        \cellcolor{green!20}\shortstack[l]{man[197,150,420,633]\\roller skates[272,523,317,595]} &
        \cellcolor{green!20}\shortstack[l]{man[197,150,420,633]\\roller skates[272,523,317,595]} &
        \cellcolor{green!20}\shortstack[l]{man[197,150,420,633]\\roller skates[272,523,317,595]} \\
        \\
        % Row 2
        \cellcolor{green!20}\shortstack[l]{bicycle helmet[328,149,409,266]\\man[259,98,429,563]} &
        \cellcolor{green!20}\shortstack[l]{roller skates[361,524,390,585]\\man[331,174,576,589]} &
        \cellcolor{green!20}\shortstack[l]{roller skates[361,524,390,585]\\man[331,174,576,589]} &
        \cellcolor{green!20}\shortstack[l]{roller skates[361,524,390,585]\\man[331,174,576,589]} \\
        \\
        % Row 3
        \cellcolor{red!20}\shortstack[l]{bicycle helmet[90,138,156,218]\\man[0,141,195,491]} &
        \cellcolor{red!20}\shortstack[l]{roller skates[262,503,280,571]\\man[331,174,576,589]} &
        \cellcolor{green!20}\shortstack[l]{man[131,98,262,530]\\bicycle helmet[90,138,156,218]} &
        \cellcolor{green!20}\shortstack[l]{man[131,98,262,530]\\bicycle helmet[90,138,156,218]} \\
        \bottomrule
    \end{tabular}

    % A single, combined caption for both the image and the table
    \caption{Example 3: Qualitative comparison of top object pairs from different methods versus the Ground Truth (GT) pairs for the given image (top). The table (bottom) details these pairs. \textcolor{green!50!black}{Green} indicates a correct pair, while \textcolor{red}{red} indicates an incorrect one.}
    \label{fig:qualitative_refinement3}
\end{figure*}

To better understand the impact of our pair refinement module, we visualize object pairs selected by each refinement strategy: semantic-only, depth-only, and the fused combination of both. For each image, we also list the ground-truth object pairs from the dataset. This comparison highlights how semantic and spatial cues contribute differently to filtering, and how their combination improves the selection of meaningful object pairs for relation prediction. Fig.~\ref{fig:qualitative_refinement1},~\ref{fig:qualitative_refinement2} and~\ref{fig:qualitative_refinement3} show examples of these.

% --- Vertically stacked Figure + Table spanning the full page width ---

% --- Vertically stacked Figure + Table spanning the full page width ---

% \newpage

\begin{figure*}[t]
    \centering
    \begin{tcolorbox}[
        title={\texttt{Semantic Pair Scoring Prompt}}, 
        colback=gray!5, 
        colframe=black!50, 
        fonttitle=\bfseries, 
        top=1pt, 
        bottom=1pt, 
        boxsep=2pt,
        left=2pt
    ]
    \small\ttfamily
    You are a world-class vision-language analyst, highly specialized in understanding spatial and functional relationships between objects in visual scenes. Your role is to evaluate how likely it is that specific object pairs are engaged in meaningful physical interactions in the given image.

    \#\#\# Object Pair List:\\
    % --- START of the two-column environment ---
    \begin{multicols*}{2} 
Pair 1: book and bookcase \\ 
Pair 2: book and bottle  \\
Pair 3: book and cat  \\
Pair 4: book and chair  \\
Pair 5: book and chest of drawers \\ 
Pair 6: book and computer monitor  \\
Pair 7: book and desk  \\
Pair 8: book and drawer  \\
Pair 9: book and lamp  \\
Pair 10: book and laptop \\ 
Pair 11: book and mouse  \\
Pair 12: book and musical keyboard \\ 
Pair 13: book and poster  \\
Pair 14: book and window  \\
Pair 15: bookcase and bottle \\  
Pair 16: bookcase and cat  \\
Pair 17: bookcase and chair  \\
Pair 18: bookcase and chest of drawers \\ 
Pair 19: bookcase and computer monitor  \\
Pair 20: bookcase and desk  \\
Pair 21: bookcase and drawer  \\
Pair 22: bookcase and lamp  \\
Pair 23: bookcase and laptop  \\
Pair 24: bookcase and mouse  \\
Pair 25: bookcase and musical keyboard \\ 
Pair 26: bookcase and poster  \\
Pair 27: bookcase and window  \\
Pair 28: bottle and cat  \\
Pair 29: bottle and chair  \\
Pair 30: bottle and chest of drawers \\ 
Pair 31: bottle and computer monitor  \\
Pair 32: bottle and desk  \\
Pair 33: bottle and drawer  \\
Pair 34: bottle and lamp  \\
Pair 35: bottle and laptop  \\
Pair 36: bottle and mouse  \\
Pair 37: bottle and musical keyboard \\ 
Pair 38: bottle and poster  \\
Pair 39: bottle and window  \\
Pair 40: cat and chair  \\
Pair 41: cat and chest of drawers \\ 
Pair 42: cat and computer monitor  \\
Pair 43: cat and desk  \\
Pair 44: cat and drawer  \\
Pair 45: cat and lamp  \\
Pair 46: cat and laptop  \\
Pair 47: cat and mouse  \\
Pair 48: cat and musical keyboard \\ 
Pair 49: cat and poster  \\
Pair 50: cat and window \\
    \end{multicols*}
    % --- END of the two-column environment ---
    
   \#\#\# Task:
\\
Carefully assess each object pair listed above and determine the likelihood that they participate in a meaningful interaction within the scene. Base your assessment on how objects of those categories typically relate in physical or functional terms within real-world images.  
\\
Provide a single integer confidence score from 1 to 5 for each pair, where: \\ 
- 1 = Very Unlikely \\  
- 2 = Unlikely  \\
- 3 = Uncertain  \\
- 4 = Likely  \\
- 5 = Very Likely \\ 

\#\#\# Output Format: \\
- Do not include any object names, explanations, or extra text.  \\
- Stop after the final pair.\\  
- You must return exactly one line per pair listed above.\\  
- Use the format: Pair [index]: [score]\\ \\

\#\#\# Begin:
    \end{tcolorbox}
    \caption{The full text of the Semantic Pair Scoring Prompt.}
    \label{fig:my-prompt}
\end{figure*}

\FloatBarrier 

% \newpage

% The strip environment creates a full-width, one-column space
% switch out of two‐column once
\onecolumn
% Your tcolorbox uses your custom style and is placed inside the strip
\begin{WidePromptBox}{\protect\texttt{Relation Generation Prompt for LLaVA}}
    \small\ttfamily
   You are a vision-language expert. Given an image with pairs of objects along with their bounding box coordinates. The bounding box coordinates are defined by (X\_top\_left, Y\_top\_left, X\_bottom\_right, Y\_bottom\_right) and are normalized between 0 and 1. 
        
       \#\#\# Object Pair List\\  
Pair 1: First object: 'woman' [0.36, 0.51, 0.49, 0.87], Second object:'chair' [0.37, 0.67, 0.57, 1.0]\\
Pair 2: First object: 'woman' [0.36, 0.51, 0.49, 0.87], Second object:'chair' [0.54, 0.64, 0.71, 1.0]\\
Pair 3: First object: 'woman' [0.36, 0.51, 0.49, 0.87], Second object:'chair' [0.7, 0.63, 0.9, 0.97]\\
Pair 4: First object: 'woman' [0.36, 0.51, 0.49, 0.87], Second object:'chair' [0.82, 0.61, 1.0, 0.93]\\
Pair 5: First object: 'woman' [0.36, 0.51, 0.49, 0.87], Second object:'chair' [0.14, 0.7, 0.32, 1.0]\\
Pair 6: First object: 'woman' [0.36, 0.51, 0.49, 0.87], Second object:'table' [0.07, 0.64, 0.54, 0.74]\\
Pair 7: First object: 'woman' [0.36, 0.51, 0.49, 0.87], Second object:'table' [0.51, 0.6, 0.77, 0.71]\\
Pair 8: First object: 'woman' [0.36, 0.51, 0.49, 0.87], Second object:'chair' [0.48, 0.57, 0.55, 0.63]\\
Pair 9: First object: 'woman' [0.36, 0.51, 0.49, 0.87], Second object:'table' [0.18, 0.55, 0.53, 0.65]\\
Pair 10: First object: 'chair' [0.37, 0.67, 0.57, 1.0], Second object:'man' [0.51, 0.52, 0.69, 0.91]\\
Pair 11: First object: 'chair' [0.37, 0.67, 0.57, 1.0], Second object:'table' [0.07, 0.64, 0.54, 0.74]\\
Pair 12: First object: 'chair' [0.37, 0.67, 0.57, 1.0], Second object:'table' [0.51, 0.6, 0.77, 0.71]\\
Pair 13: First object: 'man' [0.08, 0.42, 0.19, 0.64], Second object:'chair' [0.06, 0.51, 0.1, 0.75]\\
Pair 14: First object: 'man' [0.08, 0.42, 0.19, 0.64], Second object:'chair' [0.48, 0.57, 0.55, 0.63]\\
Pair 15: First object: 'man' [0.08, 0.42, 0.19, 0.64], Second object:'table' [0.18, 0.55, 0.53, 0.65]\\
Pair 16: First object: 'man' [0.08, 0.42, 0.19, 0.64], Second object:'table' [0.52, 0.52, 0.75, 0.62]\\
Pair 17: First object: 'chair' [0.06, 0.51, 0.1, 0.75], Second object:'table' [0.18, 0.55, 0.53, 0.65]\\
Pair 18: First object: 'man' [0.51, 0.52, 0.69, 0.91], Second object:'chair' [0.54, 0.64, 0.71, 1.0]\\
Pair 19: First object: 'man' [0.51, 0.52, 0.69, 0.91], Second object:'chair' [0.7, 0.63, 0.9, 0.97]\\
Pair 20: First object: 'man' [0.51, 0.52, 0.69, 0.91], Second object:'chair' [0.82, 0.61, 1.0, 0.93]\\
Pair 21: First object: 'man' [0.51, 0.52, 0.69, 0.91], Second object:'table' [0.07, 0.64, 0.54, 0.74]\\
Pair 22: First object: 'man' [0.51, 0.52, 0.69, 0.91], Second object:'table' [0.51, 0.6, 0.77, 0.71]\\
Pair 23: First object: 'man' [0.51, 0.52, 0.69, 0.91], Second object:'chair' [0.48, 0.57, 0.55, 0.63]\\
Pair 24: First object: 'man' [0.51, 0.52, 0.69, 0.91], Second object:'table' [0.18, 0.55, 0.53, 0.65]\\
Pair 25: First object: 'man' [0.51, 0.52, 0.69, 0.91], Second object:'table' [0.52, 0.52, 0.75, 0.62]\\
Pair 26: First object: 'chair' [0.54, 0.64, 0.71, 1.0], Second object:'man' [0.71, 0.47, 0.85, 0.92]\\
Pair 27: First object: 'chair' [0.54, 0.64, 0.71, 1.0], Second object:'man' [0.84, 0.46, 0.96, 0.78]\\
Pair 28: First object: 'chair' [0.54, 0.64, 0.71, 1.0], Second object:'table' [0.07, 0.64, 0.54, 0.74]\\
Pair 29: First object: 'chair' [0.54, 0.64, 0.71, 1.0], Second object:'table' [0.51, 0.6, 0.77, 0.71]\\
Pair 30: First object: 'chair' [0.54, 0.64, 0.71, 1.0], Second object:'table' [0.18, 0.55, 0.53, 0.65]\\
Pair 31: First object: 'man' [0.71, 0.47, 0.85, 0.92], Second object:'chair' [0.7, 0.63, 0.9, 0.97]\\
Pair 32: First object: 'man' [0.71, 0.47, 0.85, 0.92], Second object:'chair' [0.82, 0.61, 1.0, 0.93]\\
Pair 33: First object: 'man' [0.71, 0.47, 0.85, 0.92], Second object:'table' [0.51, 0.6, 0.77, 0.71]\\
Pair 34: First object: 'man' [0.71, 0.47, 0.85, 0.92], Second object:'chair' [0.48, 0.57, 0.55, 0.63]\\
Pair 35: First object: 'man' [0.71, 0.47, 0.85, 0.92], Second object:'table' [0.18, 0.55, 0.53, 0.65]\\
Pair 36: First object: 'man' [0.71, 0.47, 0.85, 0.92], Second object:'table' [0.52, 0.52, 0.75, 0.62]\\
Pair 37: First object: 'chair' [0.7, 0.63, 0.9, 0.97], Second object:'man' [0.84, 0.46, 0.96, 0.78]\\
Pair 38: First object: 'chair' [0.7, 0.63, 0.9, 0.97], Second object:'table' [0.07, 0.64, 0.54, 0.74]\\
Pair 39: First object: 'chair' [0.7, 0.63, 0.9, 0.97], Second object:'table' [0.51, 0.6, 0.77, 0.71]\\
Pair 40: First object: 'chair' [0.7, 0.63, 0.9, 0.97], Second object:'table' [0.52, 0.52, 0.75, 0.62]\\
Pair 41: First object: 'man' [0.84, 0.46, 0.96, 0.78], Second object:'chair' [0.82, 0.61, 1.0, 0.93]\\
Pair 42: First object: 'man' [0.84, 0.46, 0.96, 0.78], Second object:'table' [0.51, 0.6, 0.77, 0.71]\\
Pair 43: First object: 'man' [0.84, 0.46, 0.96, 0.78], Second object:'chair' [0.48, 0.57, 0.55, 0.63]\\
Pair 44: First object: 'man' [0.84, 0.46, 0.96, 0.78], Second object:'table' [0.52, 0.52, 0.75, 0.62]\\
Pair 45: First object: 'chair' [0.82, 0.61, 1.0, 0.93], Second object:'table' [0.51, 0.6, 0.77, 0.71]\\
Pair 46: First object: 'chair' [0.14, 0.7, 0.32, 1.0], Second object:'table' [0.07, 0.64, 0.54, 0.74]\\
Pair 47: First object: 'table' [0.07, 0.64, 0.54, 0.74], Second object:'chair' [0.48, 0.57, 0.55, 0.63]\\
Pair 48: First object: 'table' [0.51, 0.6, 0.77, 0.71], Second object:'chair' [0.48, 0.57, 0.55, 0.63]\\
Pair 49: First object: 'chair' [0.48, 0.57, 0.55, 0.63], Second object:'table' [0.18, 0.55, 0.53, 0.65]\\
Pair 50: First object: 'chair' [0.48, 0.57, 0.55, 0.63], Second object:'table' [0.52, 0.52, 0.75, 0.62]

\#\#\# Output Format Instructions \\  
- Write two sentences describing their spatial relationship. \\  
- Sentence one describes how the first object is related to the second object. \\  
- Sentence two describes how the second object is related to the first object.  \\
- Use natural but concise relationships.  \\
- Do not describe properties of a single object. \\ 
- Format your answer in the following manner:  \\
  Pair [idx]: \\ 
  Sentence1:|Sentence2: \\

\#\#\# Begin:
\end{WidePromptBox}
% still in single‐column mode here, so caption will span full page width:
% \phantomsection                         % good practice for hyperref anchors
% \captionof{figure}{The full Relation Generation Prompt for the LLaVA-Next 7b model.}
% \label{fig:rel-gen-llava}

% \newpage

% The 'strip' environment makes this section span the full page width.
\begin{WidePromptBox}{\protect\texttt{Relation Generation Prompt for Qwen2-VL}}
    % You can still use a smaller font to save space
    \small \ttfamily
    
    You are a vision-language expert. Given an image with pairs of objects along with their bounding box coordinates. The bounding box coordinates are defined by (X\_top\_left, Y\_top\_left, X\_bottom\_right, Y\_bottom\_right) and are scaled between 1 and 1000.

    \#\#\# Object Pair List  \\
Pair 1: First object: 'woman' [360, 510, 490, 870], Second object: 'chair' [370, 670, 570, 1000]  \\
Pair 2: First object: 'woman' [360, 510, 490, 870], Second object: 'chair' [540, 640, 710, 1000]  \\
Pair 3: First object: 'woman' [360, 510, 490, 870], Second object: 'chair' [700, 630, 900, 970]  \\
Pair 4: First object: 'woman' [360, 510, 490, 870], Second object: 'chair' [820, 610, 1000, 930]  \\
Pair 5: First object: 'woman' [360, 510, 490, 870], Second object: 'chair' [140, 700, 320, 1000]  \\
Pair 6: First object: 'woman' [360, 510, 490, 870], Second object: 'table' [70, 640, 540, 740]  \\
Pair 7: First object: 'woman' [360, 510, 490, 870], Second object: 'table' [510, 600, 770, 710]  \\
Pair 8: First object: 'woman' [360, 510, 490, 870], Second object: 'chair' [480, 570, 550, 630]  \\
Pair 9: First object: 'woman' [360, 510, 490, 870], Second object: 'table' [180, 550, 530, 650]  \\
Pair 10: First object: 'chair' [370, 670, 570, 1000], Second object: 'man' [510, 520, 690, 910]  \\
Pair 11: First object: 'chair' [370, 670, 570, 1000], Second object: 'table' [70, 640, 540, 740]  \\
Pair 12: First object: 'chair' [370, 670, 570, 1000], Second object: 'table' [510, 600, 770, 710]  \\
Pair 13: First object: 'man' [80, 420, 190, 640], Second object: 'chair' [60, 510, 100, 750] \\ 
Pair 14: First object: 'man' [80, 420, 190, 640], Second object: 'chair' [480, 570, 550, 630]\\  
Pair 15: First object: 'man' [80, 420, 190, 640], Second object: 'table' [180, 550, 530, 650]\\  
Pair 16: First object: 'man' [80, 420, 190, 640], Second object: 'table' [520, 520, 750, 620]\\  
Pair 17: First object: 'chair' [60, 510, 100, 750], Second object: 'table' [180, 550, 530, 650]  \\
Pair 18: First object: 'man' [510, 520, 690, 910], Second object: 'chair' [540, 640, 710, 1000]  \\
Pair 19: First object: 'man' [510, 520, 690, 910], Second object: 'chair' [700, 630, 900, 970]  \\
Pair 20: First object: 'man' [510, 520, 690, 910], Second object: 'chair' [820, 610, 1000, 930]  \\
Pair 21: First object: 'man' [510, 520, 690, 910], Second object: 'table' [70, 640, 540, 740]  
Pair 22: First object: 'man' [510, 520, 690, 910], Second object: 'table' [510, 600, 770, 710]  \\
Pair 23: First object: 'man' [510, 520, 690, 910], Second object: 'chair' [480, 570, 550, 630]  \\
Pair 24: First object: 'man' [510, 520, 690, 910], Second object: 'table' [180, 550, 530, 650]  \\
Pair 25: First object: 'man' [510, 520, 690, 910], Second object: 'table' [520, 520, 750, 620]  \\
Pair 26: First object: 'chair' [540, 640, 710, 1000], Second object: 'man' [710, 470, 850, 920]  \\
Pair 27: First object: 'chair' [540, 640, 710, 1000], Second object: 'man' [840, 460, 960, 780]  \\
Pair 28: First object: 'chair' [540, 640, 710, 1000], Second object: 'table' [70, 640, 540, 740]  \\
Pair 29: First object: 'chair' [540, 640, 710, 1000], Second object: 'table' [510, 600, 770, 710]  \\
Pair 30: First object: 'chair' [540, 640, 710, 1000], Second object: 'table' [180, 550, 530, 650]  \\
Pair 31: First object: 'man' [710, 470, 850, 920], Second object: 'chair' [700, 630, 900, 970]  \\
Pair 32: First object: 'man' [710, 470, 850, 920], Second object: 'chair' [820, 610, 1000, 930]  \\
Pair 33: First object: 'man' [710, 470, 850, 920], Second object: 'table' [510, 600, 770, 710]  \\
Pair 34: First object: 'man' [710, 470, 850, 920], Second object: 'chair' [480, 570, 550, 630]  \\
Pair 35: First object: 'man' [710, 470, 850, 920], Second object: 'table' [180, 550, 530, 650]  \\
Pair 36: First object: 'man' [710, 470, 850, 920], Second object: 'table' [520, 520, 750, 620]  \\
Pair 37: First object: 'chair' [700, 630, 900, 970], Second object: 'man' [840, 460, 960, 780]  \\
Pair 38: First object: 'chair' [700, 630, 900, 970], Second object: 'table' [70, 640, 540, 740]  \\
Pair 39: First object: 'chair' [700, 630, 900, 970], Second object: 'table' [510, 600, 770, 710]  \\
Pair 40: First object: 'chair' [700, 630, 900, 970], Second object: 'table' [520, 520, 750, 620]  \\
Pair 41: First object: 'man' [840, 460, 960, 780], Second object: 'chair' [820, 610, 1000, 930]  \\
Pair 42: First object: 'man' [840, 460, 960, 780], Second object: 'table' [510, 600, 770, 710]  \\
Pair 43: First object: 'man' [840, 460, 960, 780], Second object: 'chair' [480, 570, 550, 630]  \\
Pair 44: First object: 'man' [840, 460, 960, 780], Second object: 'table' [520, 520, 750, 620]  \\
Pair 45: First object: 'chair' [820, 610, 1000, 930], Second object: 'table' [510, 600, 770, 710]  \\
Pair 46: First object: 'chair' [140, 700, 320, 1000], Second object: 'table' [70, 640, 540, 740]  \\
Pair 47: First object: 'table' [70, 640, 540, 740], Second object: 'chair' [480, 570, 550, 630]  \\
Pair 48: First object: 'table' [510, 600, 770, 710], Second object: 'chair' [480, 570, 550, 630]  \\
Pair 49: First object: 'chair' [480, 570, 550, 630], Second object: 'table' [180, 550, 530, 650]  \\
Pair 50: First object: 'chair' [480, 570, 550, 630], Second object: 'table' [520, 520, 750, 620]\\

    \#\#\# Output Instructions  \\
    % The compact itemize list is still a good idea
    \begin{itemize}[noitemsep,topsep=1pt,leftmargin=*]
        \item[-] For each pair, write two short sentences: 
        \item[-] Sentence 1: how the first object relates to the second. 
        \item[-] Sentence 2: how the second object relates to the first. 
        \item[-] Focus on spatial or functional interactions. 
        \item[-] Use this format: 
          Pair [index]:  
          Sentence1: | Sentence2:
    \end{itemize}

    \#\#\# Begin:
\end{WidePromptBox}

% switch to two‐column again
\twocolumn

\end{document}